\documentclass{article}

\usepackage[final,nonatbib]{neurips_2021}
\usepackage[utf8]{inputenc} % allow utf-8 input
\usepackage[T1]{fontenc}    % use 8-bit T1 fonts
\usepackage{hyperref}       % hyperlinks
\usepackage{url}            % simple URL typesetting
\usepackage{booktabs}       % professional-quality tables
\usepackage{amsfonts}       % blackboard math symbols
\usepackage{nicefrac}       % compact symbols for 1/2, etc.
\usepackage{microtype}      % microtypography

\usepackage{xr}
\makeatletter
\newcommand{\maketitlepage}{%
    \let\thanks\@gobble
    \let\footnote\@gobble
    \if@twocolumn
      \ifnum \col@number=\@ne
        \@maketitle
      \else
        \twocolumn[\@maketitle]%
      \fi
    \else
      \@maketitle
    \fi
    \thispagestyle{empty}
}
\makeatother

\author{%
  David S.~Hippocampus\thanks{Use footnote for providing further information
    about author (webpage, alternative address)---\emph{not} for acknowledging
    funding agencies.} \\
  Department of Computer Science\\
  Cranberry-Lemon University\\
  Pittsburgh, PA 15213 \\
  \texttt{hippo@cs.cranberry-lemon.edu} \\
}

\usepackage{times}
\usepackage{epsfig}
\usepackage{graphicx}
\usepackage{amsmath}
\usepackage{amssymb}

\usepackage{microtype}
\usepackage{subcaption}
\usepackage{graphicx}
\usepackage{comment}
\usepackage{color}
\usepackage[utf8]{inputenc} % allow utf-8 input
\usepackage{url}            % simple URL typesetting
\usepackage{booktabs}       % professional-quality tables
\usepackage{amsfonts}       % blackboard math symbols
\usepackage{nicefrac}       
\usepackage{microtype}      
\usepackage{color}

\usepackage{listings}
\usepackage{xcolor,colortbl}

\definecolor{codegreen}{rgb}{0,0.6,0}
\definecolor{codegray}{rgb}{0.5,0.5,0.5}
\definecolor{codepurple}{rgb}{0.58,0,0.82}
\definecolor{backcolour}{rgb}{0.95,0.95,0.95}

\lstdefinestyle{mystyle}{
    backgroundcolor=\color{backcolour},   
    commentstyle=\color{codegreen},
    keywordstyle=\color{magenta},
    numberstyle=\tiny\color{codegray},
    stringstyle=\color{codepurple},
    basicstyle=\ttfamily\footnotesize,
    breakatwhitespace=false,         
    breaklines=true,                 
    captionpos=b,                    
    keepspaces=true,                 
    numbers=left,                    
    numbersep=5pt,                  
    showspaces=false,                
    showstringspaces=false,
    showtabs=false,                  
    tabsize=2
}

\definecolor{vvlightgray}{rgb}{0.9,0.9,0.9}
\definecolor{vlightgray}{rgb}{0.8,0.8,0.8}

\usepackage{wrapfig}
\graphicspath{ {./images/} }
\usepackage{multirow}
\usepackage{sidecap}

\newcommand{\blue}[1]{\textcolor{blue}{#1}}

\newcommand{\orange}[1]{\textcolor{orange}{#1}}

\DeclareMathOperator*{\argmax}{arg\,max}

\begin{document}

%%%%%%%%% TITLE
% \title{Characterising Generalization Under OOD Shifts in Deep Metric Learning}
\title{Characterizing Generalization under Out-Of-Distribution Shifts in Deep Metric Learning}
%\title{Assessing Generalization Performance Under Distribution Shifts in Deep Metric Learning}
% \title{Measuring The Depth of the Rift Between Deep Metric Learning Performance Under Distribution Shifts}
% Measuring The Rift Between Deep Metric Learning Performance Under Distribution Shifts 

% Finding acronymns...
% (ME)asuring (G)eneralization under (A)daptive Distribution Shifts
% Progressive Shifts for measuring Generalization
% (G)auging (OOD) (Splits) = GOODSplits
% (G)auging (OOD) Generalization in (DML) = ooDML
% (G)eneralization (U)nder (D)istribution (S)hifts = GUDS
%--------------------------------------------------------
% (G)eneralization under (OOD) shifts in (DML) = ooDML

\author{Timo Milbich$^*$\\
LMU Munich \& IWR, Heidelberg University\\
{\tt\small timo.milbich@iwr.uni-heidelberg.de}\\
\And
Karsten Roth$^{*, x}$\\
IWR, Heidelberg University\\
{\tt\small karsten.rh1@gmail.com}\\
\And
Samarth Sinha\\
University of Toronto, Vector\\
{\tt\small sinhasam@fb.com}\\
\And
Ludwig Schmidt\\
University of Washington\\
{\tt\small schmidt@cs.uw.edu}\\
\And
Marzyeh Ghassemi$^\dagger$\\
MIT, University of Toronto, Vector\\
{\tt\small mghassem@mit.edu}\\
\And
Bj\"orn Ommer$^\dagger$\\
LMU Munich \& IWR, Heidelberg University\\
{\tt\small ommer@uni-heidelberg.de}
}

\let\svthefootnote\thefootnote
\newcommand\freefootnote[1]{%
  \let\thefootnote\relax%
  \footnotetext{#1}%
  \let\thefootnote\svthefootnote%
}

% \author{Timo Milbich$^1$, Karsten Roth$^{1,2,4}$, Samarth Sinha$^{2}$,\\ \textbf{Ludwig Schmidt}$^{3}$, \textbf{Marzyeh Ghassemi}$^{*, 2}$, \textbf{Bjoern Ommer}$^{*, 1}$\\
% \small{$^1$IWR, Heidelberg University $^2$University of Toronto, Vector}\\ \small{$^3$University of Washington $^4$University of Tuebingen}\\
% \small{Primary contact: \url{karsten.rh1@gmail.com}}\\
% $^*$ Equal supervision
% }

\maketitle

%%%%%%%%% ABSTRACT
\begin{abstract}
Deep Metric Learning (DML) aims to find representations suitable for zero-shot transfer to \textit{a priori} unknown test distributions. However, common evaluation protocols only test a single, fixed data split in which train and test classes are assigned randomly. More realistic evaluations should consider a broad spectrum of distribution shifts with potentially varying degree and difficulty.
In this work, we systematically construct train-test splits of increasing difficulty and present the \emph{\textbf{ooDML}} benchmark to characterize generalization under \emph{\textbf{o}}ut-\emph{\textbf{o}}f-distribution shifts in \emph{\textbf{DML}}. \emph{ooDML} is designed to probe the generalization performance on much more challenging, diverse train-to-test distribution shifts. Based on our new benchmark, we conduct a thorough empirical analysis of state-of-the-art DML methods. We find that while generalization tends to consistently degrade with difficulty, some methods are better at retaining performance as the distribution shift increases. Finally, we propose few-shot DML as an efficient way to consistently improve generalization in response to unknown test shifts presented in \emph{ooDML}\footnote{Code available here: https://github.com/CompVis/Characterizing\_Generalization\_in\_DML}.
\freefootnote{$^*$ Equal contribution, alphabetical order, $^\dagger$ equal supervision, $^x$ now at University of Tuebingen.}

\end{abstract}

%However, generalizing to in-distribution samples is significantly easier than generalizing to such OOD (OOD) samples since it requires the model to extrapolate from the training samples~\cite{ood_1,ood_2,ood_3,ood_4}.
% been a successful in learning metric representation spaces suitable for zero-shot transfer to \textit{a priori} unknown test distributions. However, common evaluation protocols only test single, fixed data splits in which train and test classes are assigned arbitrarily. More realistic evaluations should consider a broad spectrum of data distribution shifts of potentially varying degree and difficulty.
%In this work, we first present a new set of train-test splits of increasing difficulty, designed to probe the generalization performance of realistic train-to-test distribution shifts. Second, we conduct a thorough empirical analysis of state-of-the-art DML performance. While generalization degrades nearly monotonous with difficulty, we find that some methods degrade more prominently as the distribution shift increases. Finally, we propose few-shot DML as an efficient way to consistently improve generalization in response to unknown test shifts. 

%%%%%%%%% BODY TEXT
\section{Introduction}\label{sec:introduction}
Image representations that generalize well are the foundation of numerous computer vision tasks, such as image and video retrieval~\cite{npairs,margin,roth2020revisiting,milbich2020diva,Brattoli_2020_CVPR}, face (re-)identification~\cite{semihard,sphereface,arcface} and image classification~\cite{tian2019contrastive,chen2020simple,moco,pretextmisra,milbich2020reliablerelations}. 
Ideally, these representations should not only capture data within the training distribution, but also transfer to new, out-of-distribution (OOD) data. However, in practice, achieving effective OOD generalization is more challenging than in-distribution~\cite{koh2021wilds,ood_1,ood_2,ood_3,ood_4,ood_5}. 
In the case of zero-shot generalization, where train and test classes are completely distinct, Deep Metric Learning (DML) is used to learn metric representation spaces that capture and transfer visual similarity to unseen classes, constituting \textit{a priori} unknown test distributions with unspecified shift. 
To approximate such a setting, current DML benchmarks use \textit{single, predefined and fixed} data splits of disjoint train and test classes, which are assigned arbitrarily
%\footnote{E.g. first and second half of alphabetically sorted classnames in CUB200-2011\cite{cub200-2011} and CARS196\cite{cars196}.}
~\cite{margin,arcface,npairs,horde,daml,dvml,hardness-aware,mic,roth2020revisiting,musgrave2020metric,kim2020proxy,proxypp,intrabatch}.
This means that \textit{(i)} generalization is only evaluated on a fixed problem difficulty, \textit{(ii)} generalization difficulty is only implicitly defined by the arbitrary data split, \textit{(iii)} the distribution shift is not measured and \textit{(iv)} cannot be not changed. 
As a result, proposed models can overfit to these singular evaluation settings,
% , e.g. by means of implicit hyperparameter overfitting \cite{roth2020revisiting} 
which puts into question the true zero-shot generalization capabilities of proposed DML models. 

In this work, we first construct a new benchmark \emph{ooDML} to characterize \emph{g}eneralization under \emph{o}ut-\emph{o}f-distribution shifts in \emph{DML}. We systematically build \emph{ooDML} as a comprehensive benchmark for evaluating OOD generalization in changing zero-shot learning settings which covers a much larger variety of zero-shot transfer learning scenarios potentially encountered in practice. 
We systematically construct training and testing data splits of increasing difficulty as measured by their Frechet-Inception Distance~\cite{fid} and extensively evaluate the performance of current DML approaches.

Our experiments reveal that the standard evaluation splits are often close to i.i.d. evaluation settings. In contrast, our novel benchmark continually evaluates models on significantly harder learning problems, providing a more complete perspective into OOD generalization in DML. 
Second, we perform a large-scale study of representative DML methods on \emph{ooDML}, and study the actual benefit of underlying regularizations such as self-supervision \cite{milbich2020diva}, knowledge distillation \cite{roth2020s2sd}, adversarial regularization~\cite{sinha2020uniform} and specialized objective functions \cite{margin,multisimilarity,arcface,kim2020proxy,roth2020revisiting}. We find that conceptual differences between DML approaches play a more significant role as the distribution shift to the test split becomes harder. \\ 
Finally, we present a study on few-shot DML as a simple extension to achieve systematic and consistent OOD generalization. As the transfer learning problem becomes harder, even very little in-domain knowledge effectively helps to adjust learned metric representation spaces to novel test distributions.
We publish our code and train-test splits on three established benchmark sets, CUB200-2011~\cite{cub200-2011}, CARS196~\cite{cars196} and Stanford Online Products (SOP)~\cite{lifted}. Similarly, we provide training and evaluation episodes for further research into few-shot DML. Overall, our contributions can be summarized as:
{
\begin{itemize}
  \setlength{\itemsep}{1pt}
  \setlength{\parskip}{0pt}
  \setlength{\parsep}{0pt}
\item Proposing the \emph{ooDML} benchmark to create a set of more realistic train-test splits that evaluate DML generalization capabilities under increasingly more difficult zero-shot learning tasks.
\item Analyzing the current DML method landscape under \emph{ooDML} to characterize benefits and drawbacks of different conceptual approaches to DML.
%\item Analyzing different methods to help understand the reasons for better OOD generalization performance by studying different factors such as model architecture and pretraining approaches.
\item Introducing and examining few-shot DML as a potential remedy for systematically improved OOD generalization, especially when moving to larger train-test distribution shifts. 
\end{itemize}
}
\section{Related Work}
% \paragraph{Deep Metric Learning (DML)} 
DML has become essential for many applications, especially in zero-shot image and video retrieval \cite{npairs,margin,mic,horde,Brattoli_2020_CVPR,milbich2017unsupervised}.
%to face verification \cite{semihard,face_verfication_inthewild,sphereface,arcface} and contrastive representation learning \cite{moco,chen2020simple,pretextmisra} for other downstream transfer usage or improved supervised classification tasks \cite{khosla2020supervised}.
Proposed approaches most commonly rely on a surrogate ranking task over tuples during training \cite{surez2018tutorial}, ranging from simple pairs \cite{contrastive} and triplets \cite{semihard} to higher-order quadruplets \cite{quadtruplet} and more generic n-tuples \cite{npairs,lifted,genlifted,multisimilarity}. These ranking tasks can also leverage additional context such as geometrical embedding structures \cite{angular,arcface}. However, due to the exponentially increased complexity of tuple sampling spaces, these methods are usually also combined with tuple sampling objectives, relying on predefined or learned heuristics to avoid training over tuples that are too easy or too hard \cite{semihard,epshn} or reducing tuple redudancy encountered during training \cite{margin,htl,smartmining,roth2020pads}.
More recent work has tackled sampling complexity through the usage of proxy-representations utilized as sample stand-ins during training, following a NCA \cite{nca} objective \cite{proxynca,kim2020proxy,proxypp}, leveraging softmax-style training through class-proxies \cite{arcface,zhai2018classification} or simulating intraclass structures \cite{softriple}.

Unfortunately, the true benefit of these proposed objectives has been put into question recently, with \cite{roth2020revisiting} and \cite{musgrave2020metric} highlighting high levels of performance saturation of these discriminative DML objectives on default benchmark splits under fair comparison.
Instead, orthogonal work extending the standard DML training paradigm through multi-task approaches \cite{Sanakoyeu_2019_CVPR,mic,milbich2020sharing}, boosting \cite{bier,abier}, attention \cite{abe}, sample generation \cite{daml,dvml,hardness-aware}, multi-feature learning \cite{milbich2020diva} or self-distillation \cite{roth2020s2sd} have shown more promise with strong relative improvements under fair comparison \cite{roth2020revisiting,milbich2020diva}, however still only in single split benchmark settings. 
It thus remains unclear how well these methods generalize in more realistic settings \cite{koh2021wilds} under potentially much more challenging, different train-to-test distribution shifts, which we investigate in this work.

% \paragraph{OOD Generalization} 
% \Karsten{Needed? Would essentially cover work e.g. utilized in \cite{ood_4}}
% \Sam{I dont think its needed. i think just the DML section is good}
%\paragraph{Generalization in Deep Metric Learning}
%\red{Not sure if a separate section is needed here.}

%%%%%%%%%%%%%%%%%%%%%%%%%%%%%%%%%%%%%%%%%%%%%%%%%%%
%%%%%%%%%%%%%%%%%%%%%%%%%%%%%%%%%%%%%%%%%%%%%%%%%%%
%\section{The \emph{ooDML} Benchmark }\label{sec:gen_study}
\section{\emph{ooDML}: Constructing a Benchmark for OOD Generalization in DML}\label{sec:gen_study}
An image representation $\phi(x)$ learned on samples $x \in \mathcal{X}_\text{train}$ drawn from some training distribution generalizes well if can transfer to test data $\mathcal{X}_\text{test}$ that are not observed during training. In the particular case of OOD generalization, the learned representation $\phi$ is supposed to transfer to samples $\mathcal{X}_\text{test}$ which are not independently and identically distributed (i.i.d.) to $\mathcal{X}_\text{train}$. A successful approach to learning such representations is DML, which is evaluated for the special case of zero-shot generalization, i.e. the transfer of $\phi$ to distributions of unknown classes~\cite{semihard,margin,horde,arcface,roth2020revisiting,musgrave2020metric}.
%%%%%%%%%%%%%%%%%%%%%%%%%%%
DML models aim to learn an embedding $\phi$ mapping datapoints $x$ into an embedding space $\Phi$, which allows to measure similarity between $x_i$ and $x_j$ as $g(\phi(x_i), \phi(x_j))$. Typically, $g$ is a predefined metric, such as the Euclidean or Cosine distance and $\phi$ is parameterized by a deep neural network. 

In realistic zero-shot learning scenarios, test distributions are not specified a priori. Thus, their respective distribution shifts relative to the training, which indicates the difficulty of the transfer learning problem, is unknown as well. To determine the generalization capabilities of $\phi$, we would ideally measure its performance on different test distributions covering a large spectrum of distribution shifts, which we will also refer to as ``problem difficulties" in this work. Unfortunately, standard evaluation protocols test the generalization of $\phi$ on a \textit{single and fixed} train-test data split of predetermined difficulty, hence only allow for limited conclusions about zero-shot generalization. 

To thoroughly assess and compare zero-shot generalization of DML models, we aim to build an evaluation protocol that resembles the undetermined nature of the transfer learning problem. In order to achieve this, we need to be able to \textit{change}, \textit{measure} and \textit{control} the difficulty of train-test data splits. 
To this end, we present an approach to construct multiple train-test splits of measurably increasing difficulty to investigate \textit{\textbf{o}}ut-\textit{\textbf{o}}f-distribution generalization in \textit{\textbf{DML}}, which make up the \textit{ooDML} benchmark. 
Our generated train-test splits resort to the established DML benchmark sets, and are subsequently used in Sec.~\ref{sec:state_of_generalization} to thoroughly analyze the current state-of-the-art in DML. 
For future research, this approach is also easily applicable to other datasets and transfer learning problems.

%%%%%%%%%%%%%%%%%
\subsection{Measuring the gap between train and test distributions}
\label{sec:measure_split_diff}
To create our train-test data splits, we need a way of measuring the distance between image datasets. This is a difficult task due to high dimensionality and natural noise in the images. 
Recently, Frechet Inception Distance (FID) \cite{fid} was proposed to measure the distance between two image distributions by using the neural embeddings of an Inception-v3 network trained for classification on the ImageNet dataset. FID assumes that the embeddings of the penultimate layer follow a Gaussian distribution, with a given mean $\mu_{\mathcal{X}}$ and covariance $\Sigma_{\mathcal{X}}$ for a distribution of images $\mathcal{X}$.
The FID between two data distributions $\mathcal{X}_1$ and $\mathcal{X}_2$ is defined as:
\begin{equation}
d(\mathcal{X}_1,\mathcal{X}_2) \triangleq \left\Vert\mu_{\mathcal{X}_1}-\mu_{\mathcal{X}_2}\right\Vert_2^2 + \text{Tr}(\Sigma_{\mathcal{X}_1}+\Sigma_{\mathcal{X}_2}-2(\Sigma_{\mathcal{X}_1}\Sigma_{\mathcal{X}_2})^\frac{1}{2}) \; ,
\label{eq:fid}
\end{equation}
In this paper, instead of the Inception network, we use the embeddings of a ResNet-50 classifier (Frechet \textit{ResNet} Distance) for consistency with most DML studies (see e.g. \cite{margin,proxypp,kim2020proxy,Sanakoyeu_2019_CVPR,mic,milbich2020diva,roth2020revisiting,intrabatch}). For simplicity, in the following sections we will still use the abbreviation \textit{FID}. 
%and \textit{continue to refer to our distribution similarity metric using a ResNet-50 backbone as FID}. 

%%%%%%%%%%%%%%%%%%%
\subsection{On the issue with default train-test splits in DML}
\begin{table}[t]
    \caption{FID scores between i.i.d. subsampled training and test sets in comparison to FID scores measured on default splits used in standard DML evaluation protocols. As can be seen, the train-test distribution shift of two out of three benchmarks are actually close i.i.d. settings, in particular when compared to the train-test splits evaluated in Fig.~\ref{fig:progression_class_swap} reaching scores over 200.}
    \centering
    \resizebox{0.8\textwidth}{!}{   
    \begin{tabular}{l||c|c|c}
         \toprule
         \textbf{Dataset $\rightarrow$} & CUB & CARS & SOP \\
         \midrule
         \textbf{Default} - different classes train/test & 52.62 & 8.59 & 3.43 \\         
         \textbf{i.i.d.} - same classes train/test & 4.87 $\pm$ 0.05 & 2.33 $\pm$ 0.03 & 0.98 $\pm$ 0.01 \\
         \bottomrule
    \end{tabular}}
\label{tab:fid_reference}    
\end{table}

To motivate the need for more comprehensive OOD evaluation protocols, we look at the split difficulty as measured by FID of typically used train-test splits and compare to i.i.d. sampling of training and test sets from the same benchmark. Empirical results in Tab.~\ref{tab:fid_reference} show that commonly utilized DML train-test splits are very close to in-distribution learning problems when compared to more out-of-distribution splits in CARS196 and SOP (see Fig.~\ref{fig:progression_class_swap}). 
This indicates that semantic differences due to disjoint train and test classes, do not necessarily relate to actual significant distribution shifts between the train and test set.
This also explains the consistently lower zero-shot retrieval performance on CUB200-2011 as compared to both CARS196 and SOP in literature~\cite{margin,multisimilarity,horde,roth2020revisiting,musgrave2020metric,milbich2020diva}, despite SOP containing significantly more classes with fewer examples per class.
In addition to the previously discussed issues of DML evaluation protocols, this further questions conclusions drawn from these protocols about the OOD generalization of representations $\phi$.
%As such, the next section will introduce our proposed method to extend given benchmark datasets into sequences of progressively harder train-test splits to approximate a range of possible distribution shifts at test time.

%%%%%%%%%%%%%%%%%%%
\subsection{Creating train-test splits of increasing difficulty}
\begin{figure*}[t]
    \centering
    \includegraphics[width=1\textwidth]{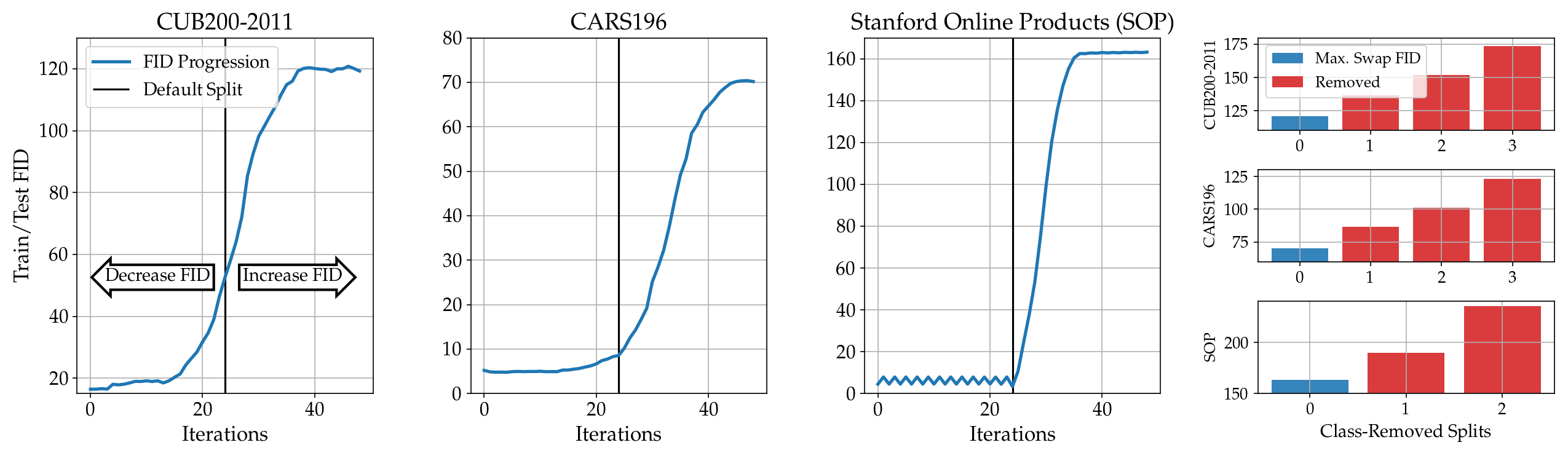}
    \caption{\textit{FID progression with iterative class swapping and removal for train-test split generation.} (Col. 1-3) FID per swapping iteration $t$ on all benchmarks.
    %With the exception for CUB200-2011, none of the default splits (black line) could be made significantly more i.i.d., highlighting the general problem of default benchmark splits.
    (Rightmost) FID of data splits obtained by additional $k$ iterations of removing classes. The blue bar denotes the maximal FID after swapping.}
    \label{fig:progression_class_swap}
\end{figure*}

Let $\mathcal{X}_\text{train}$ and $\mathcal{X}_\text{test}$ denote the original train and test set of a given benchmark dataset $\mathcal{D} = \mathcal{X}_\text{train} \cup \mathcal{X}_\text{test}$. To generate train-test splits of increasing difficulty while retaining the available data $\mathcal{D}$ and maintaining balance of their sizes, we exchange samples between them. 
To ensure and emphasize semantic consistency and unbiased data distributions with respect to image context unrelated to the target object categories, we swap entire classes instead of individual samples. Measuring distribution similarity based on FID, the goal is then to identify classes $C_\text{train} \subset \mathcal{X}_\text{train}$ and $C_\text{test} \subset \mathcal{X}_\text{test}$ whose exchange yields higher FID $d(\mathcal{X}_\text{train},\mathcal{X}_\text{test})$. 
To this end, similar to other works~\cite{dvml,mic,milbich2020diva}, we find resorting to an \textit{unimodal approximation} of the intraclass distributions sufficient and approximate FID by only considering the class means and neglect the covariance in Eq.~\ref{eq:fid}.
We select $C_\text{train}$  and $C_\text{test}$ as 
%
%\begin{align}
%    C_{tr|test}^* &= \argmax_{C_{tr|test} \in \mathcal{X}_{tr|test}} \|\mu_{C_{tr|test}} - \mu_{\mathcal{X}_{tr|test}}\|_2 - \|\mu_{C_{tr|test}} - \mu_{\mathcal{X}_{test|tr}}\|_2\label{eq:tr_select}
%\end{align}
\begin{align}
     C_\text{train}^* &= \argmax_{C_\text{train} \in \mathcal{X}_\text{train}} \|\mu_{C_\text{train}} - \mu_{\mathcal{X}_\text{train}}\|_2 - \|\mu_{C_\text{train}} - \mu_{\mathcal{X}_\text{test}}\|_2\label{eq:tr_select}\\
     C_\text{test}^* &= \argmax_{C_\text{test} \in \mathcal{X}_\text{test}} \|\mu_{C_\text{test}} - \mu_{\mathcal{X}_\text{test}}\|_2 - \|\mu_{C_\text{test}} - \mu_{\mathcal{X}_\text{train}}\|_2
     \label{eq:test_select}
\end{align}
where we measure distance to mean class-representations $\mu_{\mathcal{X}_{C}}$. By iteratively exchanging classes between data splits, i.e. $\mathcal{X}_\text{train}^{t+1} = (\mathcal{X}_\text{train}^t \setminus  C_\text{train}^*) \cup C_\text{test}^*$ and vice versa, we obtain a more difficult train-test split $(\mathcal{X}_\text{train}^{t+1},\mathcal{X}_\text{test}^{t+1})$ at iteration step $t$. 
Hence, we obtain a sequence of train-test splits $\boldsymbol{\mathcal{X}}_{\mathcal{D}} = ((\mathcal{X}_\text{train}^0,\mathcal{X}_\text{test}^0), \dots, (\mathcal{X}_\text{train}^t,\mathcal{X}_\text{test}^t), \dots,(\mathcal{X}_\text{train}^T,\mathcal{X}_\text{test}^T))$, with $\mathcal{X}_\text{train}^0 \triangleq \mathcal{X}_\text{train}$ and $\mathcal{X}_\text{test}^0 \triangleq \mathcal{X}_\text{test}$. Fig. \ref{fig:progression_class_swap} (columns 1-3) indeed shows that our FID approximation yields data splits with gradually increasing approximate FID scores with each swap until the scores cannot be further increased by swapping classes. 

UMAP visualizations in the supplementary verify that the increase corresponds to larger OOD shifts. For CUB200-2011 and CARS196, we swap two classes per iteration, while for Stanford Online Products we swap 1000 classes due to a significantly higher class count. Moreover, to cover the overall spectrum of distribution shifts and ensure comparability between benchmarks we also reverse the iteration procedure on CUB200-2011 to generate splits minimizing the approximate FID while still maintaining disjunct train and test classes.
%Fig. \ref{fig:progression_class_swap} shows this process also on the other two benchmarks with early saturation, highlighting that default splits can not be made significantly easier.

To further increase $d(\mathcal{X}_\text{train}^{T},\mathcal{X}_\text{test}^{T})$ beyond convergence (see Fig. \ref{fig:progression_class_swap}) of the swapping procedure, we subsequently also identify and remove classes from both $\mathcal{X}_\text{train}^{T}$ and  $\mathcal{X}_\text{test}^{T}$. More specifically, we remove classes $C_\text{train}$ from $\mathcal{X}_\text{train}^{T}$ that are closest to the mean of $\mathcal{X}_\text{test}^{T}$ and vice versa. For $k$ steps, we successively repeat class removal as long as $50\%$ of the original data is still maintained in these additional train-test splits.
%Hence, from a train set $\mathcal{X}_\text{train}^{T}$ we remove
%
% \begin{equation}\label{eq:tr_select_removed}
%     C_\text{train}^* = \argmax_{C_\text{train} \in \mathcal{X}_\text{train}} \|\mu_{C_\text{train}} - \mu_{\mathcal{X}_\text{test}}\|_2
% \end{equation}
 %
 %and analogously identify and remove classes $C_\text{test}^*$ from $\mathcal{X}_\text{test}^{T}$,
 %
% \begin{equation}\label{eq:test_select_removed}
%     C_\text{test}^* = \argmax_{C_\text{test} \in \mathcal{X}_\text{test}} \|\mu_{C_\text{test}} - \mu_{\mathcal{X}_\text{train}}\|_2
% \end{equation}
 %
Fig.~\ref{fig:progression_class_swap} (rightmost) shows how splits generated through class removal progressively increase the approximate FID beyond what was achieved only by swapping. To analyze if the generated data splits are not inherently biased to the used backbone network for FID computation, we also repeat this procedure based on representations from different architectures, pretraining methods and datasets in the supplementary. Note, that comparison of absolute FID values between datasets may not be meaningful and we are mainly interested in distribution shifts within a given dataset distribution.
Overall, using class swapping and removal we select splits that cover the broadest FID range possible, while still maintaining sufficient data. Hence, our splits are significantly harder and more diverse than the default splits.

%%%%%%%%%%%%%%%%%%%%%%%%%%%%%%%%%%%%%%%%%%%%%%%%%%%
%%%%%%%%%%%%%%%%%%%%%%%%%%%%%%%%%%%%%%%%%%%%%%%%%%%
\section{Assessing the State of Generalization in Deep Metric Learning}\label{sec:state_of_generalization}
\begin{figure*}[t]
    \centering
    \includegraphics[width=1\textwidth]{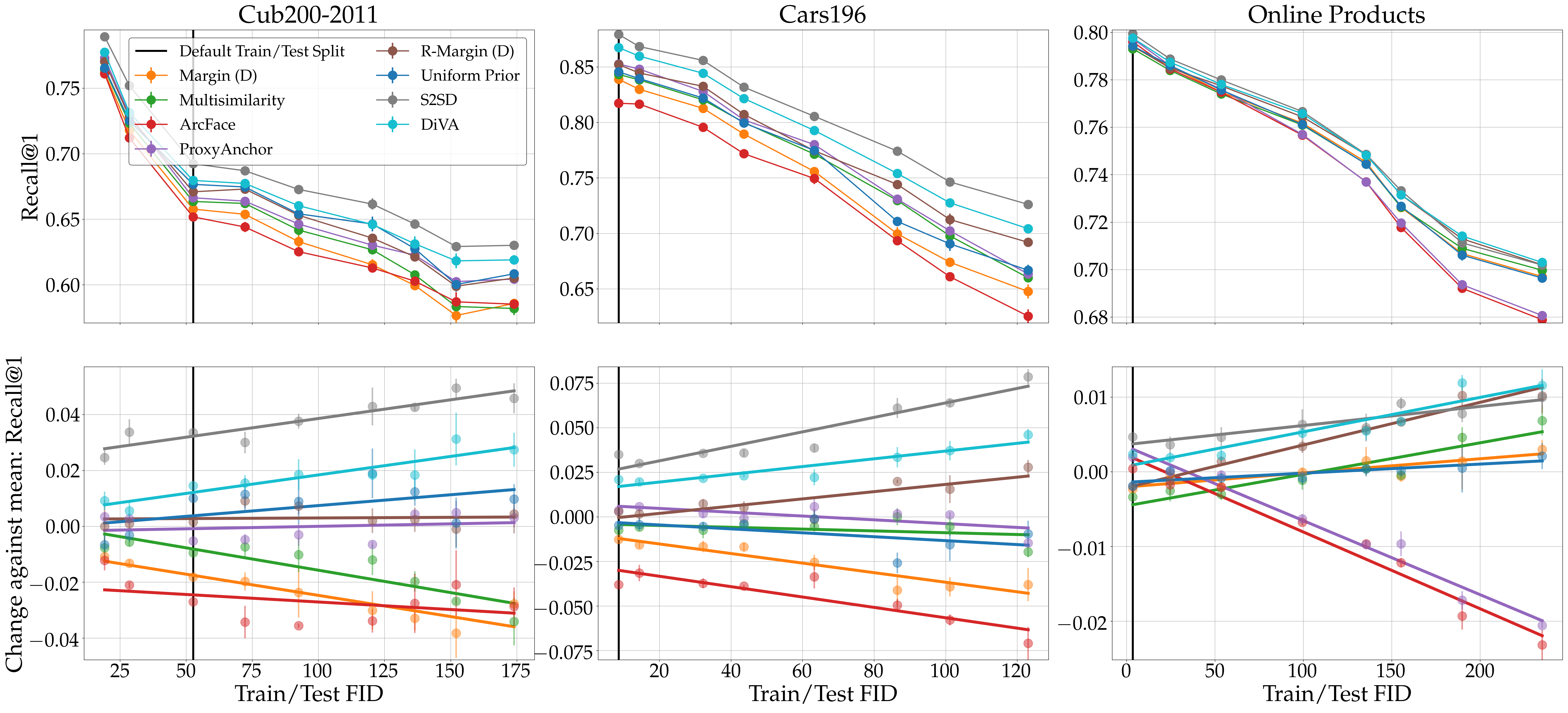}
    \caption{\textit{Zero-Shot Generalization performance under varying distribution shifts.} (top row) Absolute Recall@1 performance for each increasingly more difficult train-test split in the \textit{goodDML} benchmark (cf. Sec.~\ref{sec:gen_study}) on CUB200-2011, CARS196 and SOP.
    We report mean Recall@1 performances and standard deviations over 5 runs. For results based on mAP@1000 see supplementary. (bottom) Differences of performances against the mean over all methods for each train-test split.}
    \label{fig:total_shift_comparison}
\end{figure*}

This section assesses the state of zero-shot generalization in DML via a large experimental study of representative DML methods on our \textit{ooDML} benchmark, offering a much more complete and thorough perspective on zero-shot generalization in DML as compared to previous DML studies~\cite{fehervari2019unbiased,roth2020revisiting,musgrave2020metric,milbich2020sharing}.

For our experiments we use the three most widely used benchmarks in DML, CUB200-2011\cite{cub200-2011}, CARS196\cite{cars196} and Stanford Online Products\cite{lifted}. For a complete list of implementation and training details see the supplementary if not explicitly stated in the respective sections. Moreover, to measure generalization performance, we resort to the most widely used metric for image retrieval in DML, Recall@k~\cite{recall}. Additionally, we also evaluate results over mean average precision (mAP@1000)~\cite{roth2020revisiting,musgrave2020metric}, but provide respective tables and visualizations in the supplementary when the interpretation of results is not impacted.
%To generate all splits, feature representations extracted from the penultimate layer of a ResNet50 \cite{resnet} network pretrained on ImageNet was used, the most commonly evaluated backbone architectures both in DML (as noted in Sec. \ref{sec:measure_split_diff}) and representation learning in general (e.g. \cite{moco,chen2020simple,pretextmisra}).

The exact training and test splits ultimately utilized throughout this work are selected based on Fig.~\ref{fig:progression_class_swap} to ensure approximately uniform coverage of the spectrum of distribution shifts within intervals ranging from the lowest (near i.i.d. splits) to the highest generated shift achieved with class removal. For experiments on CARS196 and Stanford Online Products, eight total splits were investigated, included the original default benchmark split. For CUB200-2011, we select nine splits to also account for benchmark additions with reduced distributional shifts. The exact FID ranges are provided in the supplementary. 
Training on CARS196 and CUB200-2011 was done for a maximum of 200 epochs following standard training protocols utilized in \cite{roth2020revisiting}, while 150 epochs were used for the much larger SOP dataset. Additional training details if not directly stated in the respective sections can be found in the supplementary.

%%%%%%%%%%%%%%%%%%%%%%%%
\subsection{Zero-shot generalization under varying distribution shifts}\label{subsec:base_exp_results}

\begin{figure*}[t]
    \centering
    \includegraphics[width=1\textwidth]{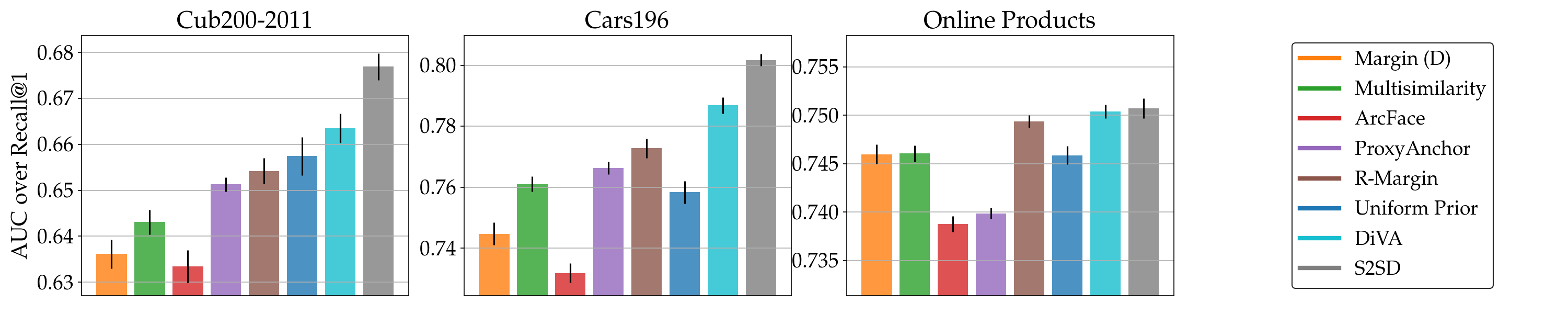}
    \caption{\textit{Comparison of DML methods via AGS based on Recall@1 across benchmarks.} To compute AGS (cf. Sec.~\ref{subsec:base_exp_results}), we aggregate the performances from Fig.~\ref{fig:total_shift_comparison} across all train-test distribution shifts of our proposed \textit{ooDML} benchmark using the Area-Under-the-Curve metric.}
    \label{fig:auc_shift_comparison}
\end{figure*}

Many different concepts have been proposed in DML to learn embedding functions $\phi$ that generalize from the training distribution to differently distributed test data. To analyze the zero-shot transfer capabilities of DML models, we consider representative approaches making use of the following concepts: 
\textit{(i)} surrogate ranking tasks and tuple mining heuristics (Margin loss with distance-based sampling \cite{margin} and Multisimilarity loss \cite{multisimilarity}), \textit{(ii)} geometric constraints or class proxies (ArcFace \cite{arcface}, ProxyAnchor \cite{kim2020proxy}), \textit{(iii)} learning of semantically diverse features (R-Margin \cite{roth2020revisiting}) and self-supervised training (DiVA~\cite{milbich2020diva}), adversarial regularization (Uniform Prior~\cite{sinha2020uniform}) and \textit{(iv)} knowledge self-distillation (S2SD~\cite{roth2020s2sd}).

Fig.~\ref{fig:total_shift_comparison} (top) analyzes these methods for their generalization to distribution shifts the varying degrees represented in \textit{ooDML}. The top row shows absolute zero-shot retrieval performance measured on Recall@1 (results for mAP@1000 can be found in the supplementary) with respect to the FID between train and test sets. Additionally, Fig.~\ref{fig:total_shift_comparison} (bottom) examines the relative differences of performance to the performance mean over all methods for each train-test split. Based on these experiments, we make the following main observations: 

\textbf{\textit{(i)} Performance deteriorates monotonically with the distribution shifts.} Independent of dataset, approach or evaluation metric, performance drops steadily as the distribution shift increases. 

\textbf{\textit{(ii)} Relative performance differences are affected by train-test split difficulty.} We see that the overall ranking between approaches oftentimes remains stable on the CARS196 and CUB200-2011 datasets. However, we also see that particularly on a large-scale dataset (SOP), established proxy-based approaches ArcFace~\cite{arcface} (which incorporates additional geometric constraints) and ProxyAnchor~\cite{kim2020proxy} are surprisingly susceptible to more difficult distribution shifts. 
Both methods perform poorly compared to the more consistent general trend of the other approaches. Hence, conclusions on the generality of methods solely based on the default benchmarks need to be handled with care, as at least for SOP, performance comparisons reported on single (e.g. the standard) data splits \textit{do not} translate to more general train-test scenarios.

\textbf{\textit{(iii)} Conceptual differences matter at larger distribution shifts} While the ranking between most methods is largely consistent on CUB200-2011 and CARS196, their differences in performance becomes more prominent with increasing distribution shifts. The relative changes (deviation from the mean of all methods at the stage) depicted in Fig.~\ref{fig:total_shift_comparison} (bottom) clearly indicates that particular methods based on machine learning techniques such as self-supervision, feature diversity (DiVA, R-Margin) and self-distillation (S2SD) are among the best at generalizing in DML on more challenging splits while retaining strong performance on more i.i.d. splits as well.
% This verifies other research results from representation learning, indicating benefits of self-supervised training for transfer learning~\cite{moco,chen2020simple}, self-distillation for generalization improvement~\cite{tian2020rethinking} as well as similar findings for zero-shot settings in the NLP domain~\cite{dalle}.
%Interestingly however, the benefit of these orthogonal extension to DML not only benefit performance at higher degrees of "OODness", but also in more i.i.d. settings as well, although differences in methods is notably less significant. These results further highlight the benefit of orthogonal extensions to the otherwise discriminative training process in DML, with auxiliary self-supervision through DiVA and self-distillation and feature reusability through S2SD providing a better starting point when test shifts are unknown beforehand.

While directly stating performance in dependence to the individual distribution shifts offers a detailed overview, the overall comparison of approaches is typically based on single benchmark scores. To further provide a single metric of comparison, we utilize the well-known Area-under-Curve (AUC) score to condense performance (either based on Recall@1 or mAP@1000) over all available distribution shifts into a single aggregated score indicating general zero-shot capabilities. This \textit{Aggregated Generalization Score (AGS)} is computed based on the normalized FID scores of our splits to the interval $[0,1]$. As Recall@k or mAP@k scores are naturally bounded to $[0,1]$, AGS is similarly bound to $[0,1]$ with higher being the better model. Our corresponding results are visualized in Fig.~\ref{fig:auc_shift_comparison}. Indeed, we see that AGS reflects our observations from Fig.~\ref{fig:total_shift_comparison}, with self-supervision (DiVA) and self-distillation (S2SD) generally performing best when facing unknown train-test shifts. Exact scores are provided in the supplementary.

%%%%%%%%%%%%%%%%%%%%%%%%
\subsection{Consistency of structural representation properties on \textit{ooDML}}
\label{sec:gen_metrics}
\begin{figure}[h]
    \centering
    \includegraphics[width=1\textwidth]{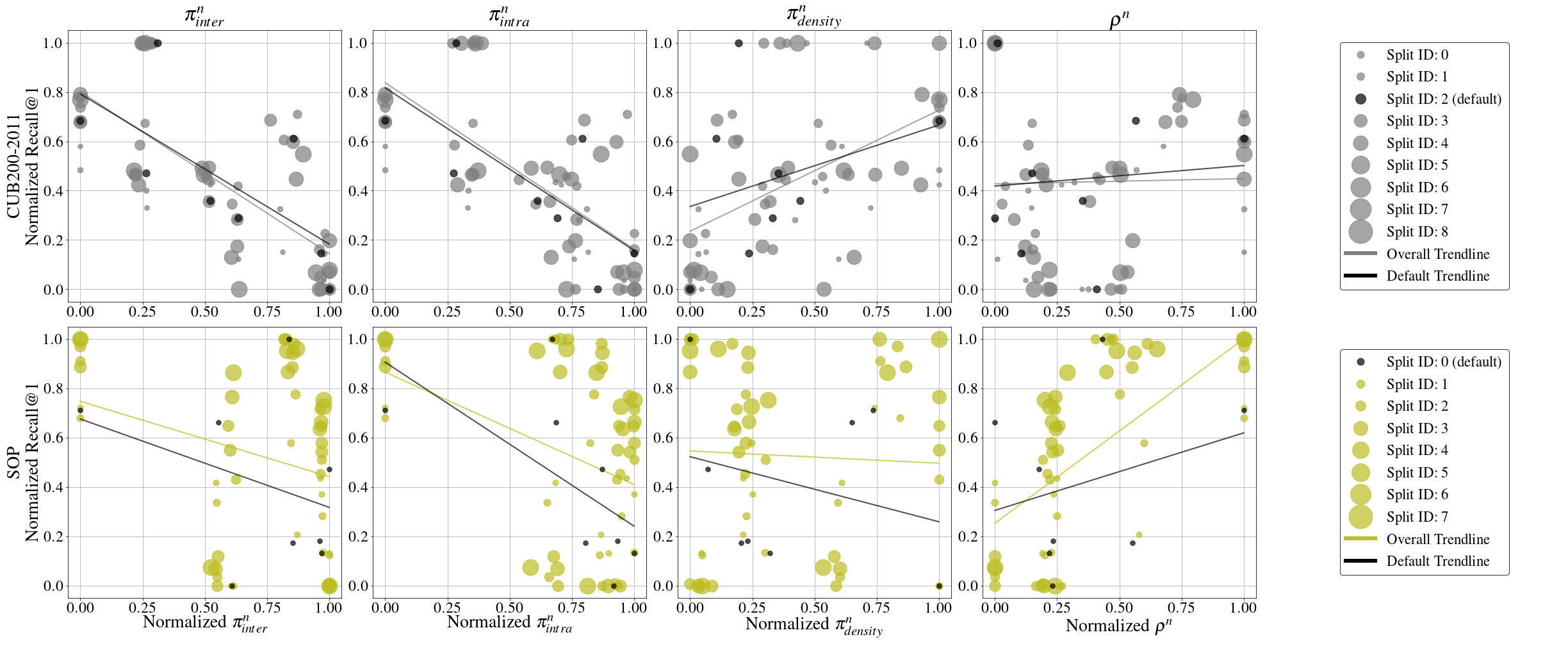}
    \caption{\textit{Generalization metrics computed on ooDML benchmark for CUB200-2011 and SOP.} Each column plots one of the (normalized) measured structural representation property (cf.~\ref{sec:gen_metrics}) over the corresponding Recall@1 performance for all examined DML methods and distribution shifts. Trendlines are computed as least squares fit over all datapoints (overall), respectively only those corresponding to default splits (default).}
    \label{fig:metrics_recall}
\end{figure}
% \begin{figure}[h]
%     \centering
%     \includegraphics[width=1\textwidth]{images/relative_metrics_map.png}
%     \caption{\textit{Generalization metrics.}}
%     \label{fig:metrics_map}
% \end{figure}

Roth et al.~\cite{roth2020revisiting} attempts to identify potential drivers of generalization in DML by measuring the following structural properties of a representation $\phi$: \textit{(i)} the mean distance $\pi_\text{inter}$ between the centers of the embedded samples of each class, \textit{(ii)} the mean distance $\pi_\text{intra}$ between the embedded samples within a class $\pi_\text{intra}$, \textit{(iii)} the `embedding space density' measured as the ratio $\pi_\text{ratio} = \frac{\pi_\text{intra}}{\pi_\text{inter}}$ and \textit{(iv)} `spectral decay' $\rho(\Phi)$ measuring the degree of uniformity of singular values obtained by singular value decomposition on the training sample representations, which indicates the number of significant directions of variance. For a more detailed description, we refer to \cite{roth2020revisiting}. These metrics indeed are empirically shown to exhibit a certain correlation to generalization performance on the default benchmark splits. In contrast, we are now interested if these observations hold when measuring generalization performance on the \textit{ooDML} train-test splits of varying difficulty.
%as well as extending the study done in \cite{roth2020revisiting} to account for regulatory extensions to DML such as via self-supervision instead of solely discriminative DML training objectives.

We visualize our results in Fig.~\ref{fig:metrics_recall} for CUB200-2011 and SOP, with CARS196 provided in the supplementary. For better visualization we normalize all measured values obtained for both metrics \textit{(i)}-\textit{(iv)} and the recall performances (Recall@1) to the interval $[0, 1]$ for each train-test split. Thus, the relation between structural properties and generalization performance becomes comparable across all train-test splits, allowing us to examine if superior generalization is still correlated to the structural properties of the learned representation $\phi$, i.e. if the correlation is independent of the underlying distribution shifts. 
For a perfectly descriptive metric, one should expect to see heavy correlation between normalized metric and normalized generalization performance jointly across shifts. Unfortunately, our results show only a small indication of any structural metric being consistently correlated with generalization performance over varying distribution shifts. 
This is also observed when evaluating only against basic, purely discriminative DML objectives as was done in \cite{roth2020revisiting} for the default split, as well as when incorporating methods that extend and change the base DML training setup (such as DiVA \cite{milbich2020diva} or adversarial regularization \cite{sinha2020uniform}).
%For the later, we also find that even on default splits (black dots) are correlations not consistent.
% This This contrasts insights from \cite{roth2020revisiting} who measure correlation only for standard, purely discriminative DML objectives. 

This not only demonstrates that experimental conclusions derived from the analysis of only single benchmark split may not hold for overall zero-shot generalization, but also that future research should consider more general learning problems and difficulty to better understand the conceptual impact various regulatory approaches. To this end, our benchmark protocol offers more comprehensive experimental ground for future studies to find potential drivers of zero-shot generalization in DML.

\subsection{Network capacity and pretrained representations}
\label{sec:network_and_pretrain}
\begin{figure*}[t]
    \centering
    \includegraphics[width=1\textwidth]{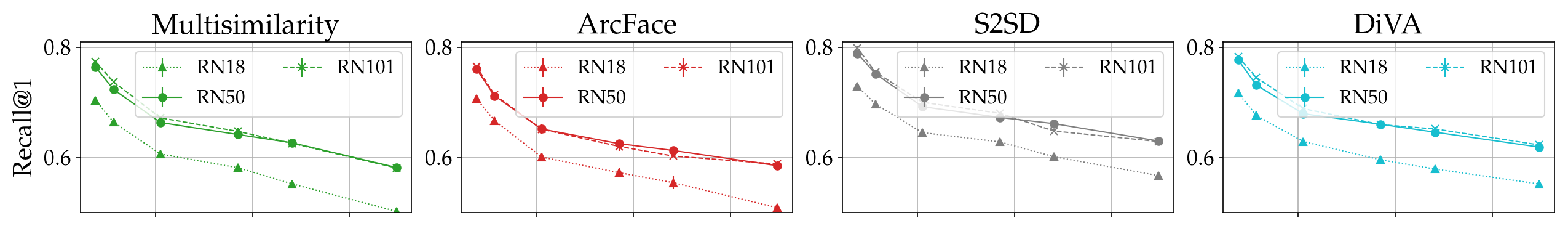}
    \caption{\textit{Generalization performance for different backbone architectures for varying distribution shifts on CUB200-2011.} We show absolute Recall@1 performances averaged over 5 runs for each train-test split. Other datasets show similar results and are provided in the supplementary.}
    \label{fig:arch_comp}
\end{figure*}

A common way to improve generalization, as also highlighted in \cite{roth2020revisiting} and \cite{musgrave2020metric}, is to select a stronger backbone architecture for feature extraction. In this section, we look at how changes in network capacity can influence OOD generalization across distribution shifts. Moreover, we also analyze the zero-shot performance of a diverse set of state-of-the-art pretraining approaches.
% As such, in this section, we look at how changes in network capacity, but also general architecture as well as pretraining method as means of incorporating predefined inductive biases into the training process, can influence OOD generalization across distribution shifts. 
%

\textbf{Influence of network capacity.} 
%We begin by examining how the overall capacity of the network, as measured by its number of weights/layers with otherwise retained architecture, can impact generalization and how this relates to changes in dataset shifts. 
In Fig.~\ref{fig:arch_comp}, we compare different members of the ResNet architecture family~\cite{resnet} with increasing capacity, each of which achieve increasingly higher performance on i.i.d. test benchmarks such as ImageNet \cite{imagenet}, going from a small ResNet18 (R18) over ResNet50 (R50) to ResNet101 (R101) variants. As can be seen, while larger network capacity helps to some extent, we observe that performance actually saturates in zero-shot transfer settings, regardless of the DML approach and dataset (in particular also the large scale SOP dataset). 
Interestingly, we also observe that the performance drops with increasing distribution shifts are consistent across network capacity, suggesting that zero-shot generalization is less driven by network capacity but rather conceptual choices of the learning formulation (compare Fig.~\ref{fig:total_shift_comparison}).
%
% And while there is a slight relative change in generalization performance, overall performance scales similarly with increasingly larger generalization gaps for all considered architectures. This verifies that OOD generalization actually is driven by different concepts than (standard discriminative) in-distribution generalization, i.e. stronger in-distribution performance does not necessarily translate to increased OOD generalization. To this end, recent DML approaches have focused on different ways to bridge the generalization gap and improve zero-shot performance, which we analyze is the next section.
%
% \textbf{Training vs. Pretraining.}

\textbf{Generic representations versus Deep Metric Learning.}
\begin{figure}[t]
    \centering
    \includegraphics[width=1\textwidth]{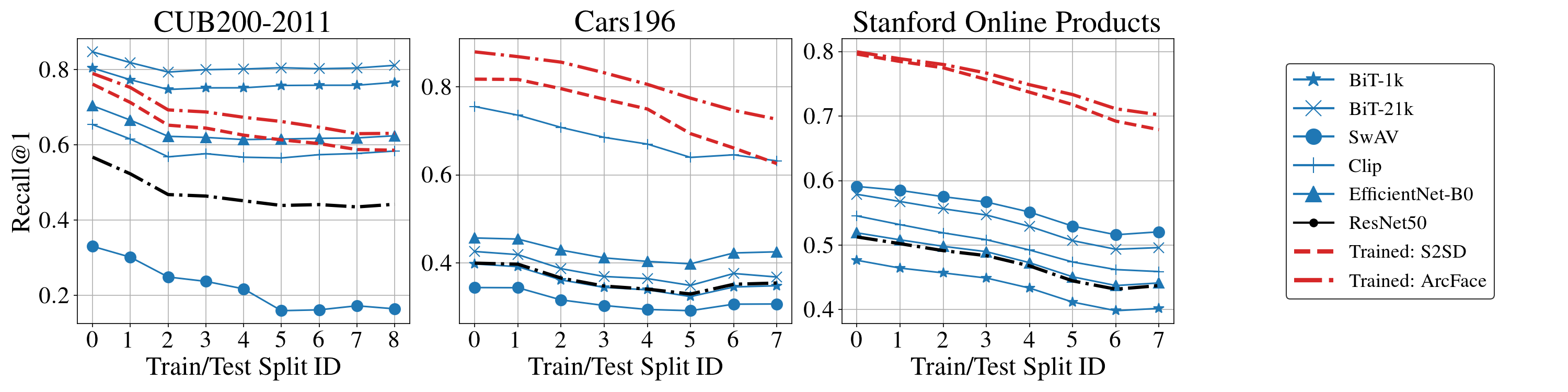}
    \caption{\textit{Comparison of DML to various non-adapted generic representations pretrained on large amounts unlabelled data and state-of-the-art architectures.} For DML, we show best and worst DML objectives based on results in Fig.~\ref{fig:total_shift_comparison}. Performance of generic representations are heavily dependent on target dataset, architecture, amount of training data and learning objective.}
    \label{fig:pretrained_zeroshot_comp}
\end{figure}
Recently, self-supervised representation learning has taken great leaps with ever stronger models trained on huge amounts of image~\cite{bit,clip} and language data~\cite{bert,roberta,GPT3}. These approaches are designed to learn expressive, well-transferring features and methods like CLIP~\cite{clip} even prove surprisingly useful for zero-shot classification. We now evaluate and compare such representations against state-of-the-art DML models to understand if generic representations that are readily available nowadays actually pose an alternative to explicit application of DML.
%We now investigate how inductive biases, either through the choice of architecture or pretraining method, are linked to generalization performance across problem difficulties and check how this compares to actual adaptation to the data at hand through learning.
We select state-of-the-art self-supervision models \textit{SwAV} \cite{swav} (ResNet50 backbone), CLIP \cite{clip} trained via natural language supervision on a large dataset of 400 million image and sentence pairs (VisionTransformer \cite{vit} backbone), BiT(-M) \cite{bit}, which trains a ResNet50-V2 \cite{bit} on both the standard ImageNet \cite{imagenet} (1 million training samples) and the ImageNet-21k dataset \cite{imagenet,im21k_check} with 14 million training samples and over 21 thousand classes, an EfficientNet-B0 \cite{efficientnet} trained on ImageNet, and a standard baseline ResNet50 network trained on ImageNet. 
%To compare against Deep Metric Learning approaches, we select the generally best and worst performing methods from section \ref{sec:state_of_generalization}.
Note, that none of these representations has been additionally adapted to the benchmark sets, in contrast to the DML approaches which have been trained on the respective train splits.

The results presented in Fig.~\ref{fig:pretrained_zeroshot_comp} show large performance differences of the pretrained representations, which are largely dependent on the test dataset. 
While BiT outperforms the DML state-of-the-art on CUB200-2011 without any finetuning, it significantly trails behind the DML models on the other two datasets. On CARS196, only CLIP comes close to the DML approaches when the distribution shift is sufficiently large. Finally, on SOP, none of these models comes even close to the adapted DML methods. This shows how although representations learned by extensive pretraining can offer strong zero-shot generalization, their performance heavily depends on the target dataset and specific model. Furthermore, the generalization abilities notably depend on the size of the pretraining dataset (compare e.g. BiT-1k vs BiT-21k or CLIP), which is significantly larger than the number of training images seen by the DML methods. We see that only actual training on these datasets provides sufficiently reliable performance. 
\subsection{Few-shot adaption boosts generalization performance in DML}\label{subsec:fewshot}
\begin{figure}[t]
    \centering
    \includegraphics[width=1\textwidth]{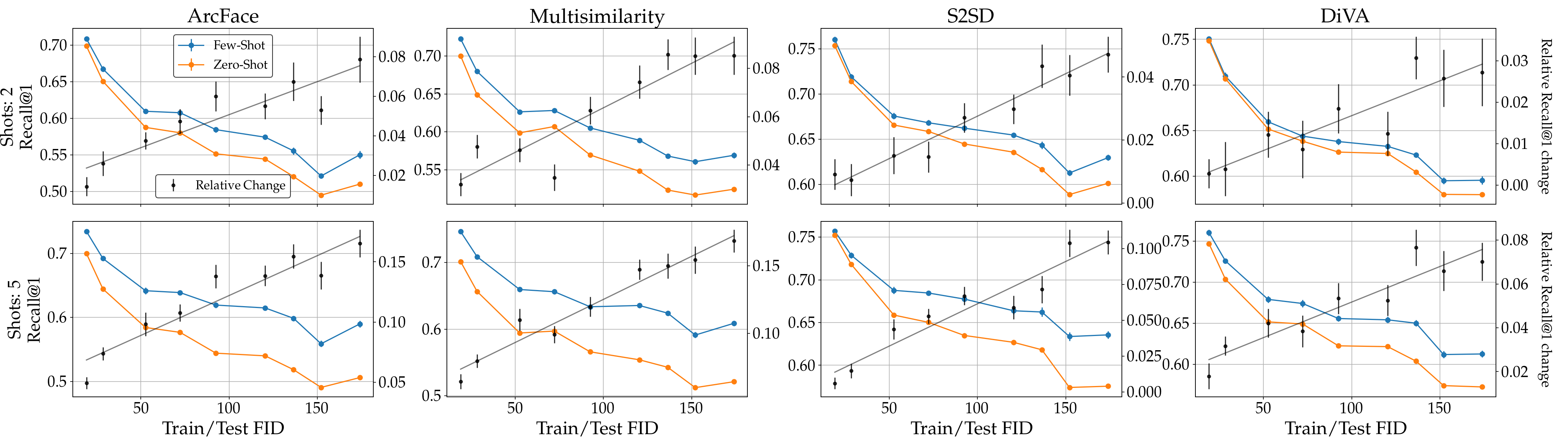}
    \caption{\textit{Few-Shot adaptation of DML representations on CUB200-2011.} Columns show average Recall@1 performance over 10 episodes of 2- and 5-shot adaption for various DML approaches (\blue{fewshot} and \orange{zeroshot}), highlighting a substantial benefit of few-shot adaptation for \textit{a priori} unknown distribution shifts (see black line highlighting relative improvements).}
    \label{fig:fewshot_cub_recall}
\end{figure}

Since distribution shifts can be arbitrarily large, the zero-shot transfer of $\phi$ can be ill-posed. Features learned on a training set $\mathcal{X}_\text{train}$ will not meaningfully transfer to test samples $\mathcal{X}_\text{test}$ once they are sufficiently far from $\mathcal{X}_\text{train}$, as also already indicated by Fig.~\ref{fig:total_shift_comparison}. As a remedy \textit{few-shot learning}~\cite{protonet,fewshot_info,maml,anil,metaoptnet,metabaseline,tian2020rethinking} assumes few samples of the test distribution to to be available during training, i.e. adjusting a previously learned representation. While these approaches are typically explicitly trained for fast adaption to novel classes, we are now interested if similar adaptation of DML representations $\phi$ helps to bridge increasingly large distribution shifts.

To investigate this hypothesis, we follow the evaluation protocol of few-shot learning and use $k$ representatives (also referred to as \textit{shots}) of each class from a test set $\mathcal{X}_\text{test}$ as a support set for finetuning the penultimate embedding network layer. The remaining test samples then represent the new test set to evaluate retrieval performance, also referred to as query set. For evaluation we perform $10$ episodes, i.e. we repeat and average the adaptation of $\phi$ over $10$ different, randomly sampled support and corresponding query sets. Independent of the DML model used for learning the original representation $\phi$ on $\mathcal{X}_\text{train}$, adaptation to the support data is conducted using the Marginloss~\cite{margin} objective with distance-based sampling \cite{margin} due to its faster convergence. This also ensures fair comparisons to the adaptation benefit to $\phi$ and also allows to adapt complex approaches like self-supervision (DiVA~\cite{milbich2020diva}) to the small number of samples in the support sets. 

Fig.~\ref{fig:fewshot_cub_recall} shows 2 and 5 shot results on CUB200-2011, with CARS196 available in the supplementary. SOP is not considered since each class is already composed of small number of samples. As we see, even very limited in-domain data can significantly improve generalization performance, with the benefit becoming stronger for larger distribution shifts. Moreover, we observe that weaker approaches like ArcFace~\cite{arcface} seem to benefit more than state-of-the-art methods like S2SD~\cite{roth2020s2sd} or DiVA~\cite{milbich2020diva}. We presume this to be caused by their underlying concepts already encouraging learning of more robust and general features. To conclude, few-shot learning provides a substantial and reliable benefit when facing OOD learning settings where the shift is not known \textit{a priori}.

\section{Conclusion}
In this work we analyzed zero-shot transfer of image representations learned by Deep Metric Learning (DML) models. We proposed a systematic construction of train-test data splits of increasing difficulty, as opposed to standard evaluation protocols that test out-of-distribution generalization only on single data splits of fixed difficulty. 
Based on this, we presented the novel benchmark \textit{ooDML} and thoroughly assessed current DML methods. Our study reveals the following main findings: 
\\
\textbf{Standard evaluation protocols are insufficient to probe general out-of-distribution transfer:} 
Prevailing train-test splits in DML are often close to i.i.d. evaluation settings. Hence, they only provide limited insights into the impact of train-test distribution shift on generalization performance. Our benchmark \textit{ooDML} alleviates this issue by evaluating a large, controllable and measurable spectrum of problem difficulty to facilitate future research.  
\\
\textbf{Larger distribution shifts show the impact of conceptual differences in DML approaches:} 
Our study reveals that generalization performance degrades consistently with increasing problem difficulty for all DML methods. However, certain concepts underlying the approaches are shown to be more robust to shifts than others, such as semantic feature diversity and knowledge-distillation.
\\
\textbf{Generic, self-supervised representations without finetuning can surpass dedicated data adaptation:}
When facing large distribution shifts, representations learned only by self-supervision on large amounts of of unlabelled data are competitive to explicit DML training without any finetuning. However, their performance is heavily dependent on the data distribution and the models themselves.
\\
\textbf{Few-shot adaptation consistently improves out-of-distribution generalization in DML:} 
Even very few examples from a target data distribution effectively help to adapt DML representations. The benefit becomes even more prominent with increasing train-test distribution shifts, and encourages further research into few-shot adaptation in DML.

%In this work we analyzed zero-shot transfer of image representations learned by Deep Metric Learning models. In contrast to prevailing evaluation protocols operating on data splits of fixed training to test distribution shifts, we assess zero-shot generalization on a much more realistic and unconstrained level. To this end, we presented a principled way to construct train-test splits of increasing difficulty. Based on this, we established a new benchmark protocol that allows us to study how generalization performance scales with the learning problem difficulty. We showed that differences in performance among DML methods get more prominent as the distribution shift increases and presented few-shot DML as an efficient way to consistently improve generalization in response to unknown test shifts. 

\paragraph{Funding transparency statement}
This research has been funded by the German Federal Ministry for Economic Affairs and Energy
within the project “KI-Absicherung – Safe AI for automated driving” and by the German Research
Foundation (DFG) within projects 371923335 and 421703927. Moreover, it was funded in part by a CIFAR AI Chair at the Vector Institute, Microsoft Research, and an NSERC Discovery Grant. Resources used in preparing this research were provided, in part, by the Province of Ontario, the Government of Canada through CIFAR, and companies sponsoring the Vector Institute \url{www.vectorinstitute.ai/#partners}.

\paragraph{Acknowledgements}
% We would like to thank Nathan Ng, Aparna Balagopalan and Hammaad Adam (University of Toronto, Vector) for insightful discussions and feedback on the paper draft.
We thank the International Max Planck Research School for Intelligent Systems (IMPRS-IS) for supporting K.R; K.R. acknowledges his membership in the European Laboratory for Learning and Intelligent Systems (ELLIS) PhD program.

%%%%%%%%%%%%%%%%%%%%%%%%%%%%%%%%%%
{\small
\bibliographystyle{ieee_fullname.bst}
\bibliography{egbib}

\begin{thebibliography}{10}\itemsep=-1pt

\bibitem{Brattoli_2020_CVPR}
Biagio Brattoli, Joseph Tighe, Fedor Zhdanov, Pietro Perona, and Krzysztof
  Chalupka.
\newblock Rethinking zero-shot video classification: End-to-end training for
  realistic applications.
\newblock In {\em Proceedings of the IEEE/CVF Conference on Computer Vision and
  Pattern Recognition (CVPR)}, June 2020.

\bibitem{GPT3}
Tom Brown, Benjamin Mann, Nick Ryder, Melanie Subbiah, Jared~D Kaplan, Prafulla
  Dhariwal, Arvind Neelakantan, Pranav Shyam, Girish Sastry, Amanda Askell,
  Sandhini Agarwal, Ariel Herbert-Voss, Gretchen Krueger, Tom Henighan, Rewon
  Child, Aditya Ramesh, Daniel Ziegler, Jeffrey Wu, Clemens Winter, Chris
  Hesse, Mark Chen, Eric Sigler, Mateusz Litwin, Scott Gray, Benjamin Chess,
  Jack Clark, Christopher Berner, Sam McCandlish, Alec Radford, Ilya Sutskever,
  and Dario Amodei.
\newblock Language models are few-shot learners.
\newblock In H. Larochelle, M. Ranzato, R. Hadsell, M.~F. Balcan, and H. Lin,
  editors, {\em Advances in Neural Information Processing Systems}, volume~33,
  pages 1877--1901. Curran Associates, Inc., 2020.

\bibitem{swav}
Mathilde Caron, Ishan Misra, Julien Mairal, Priya Goyal, Piotr Bojanowski, and
  Armand Joulin.
\newblock Unsupervised learning of visual features by contrasting cluster
  assignments.
\newblock In H. Larochelle, M. Ranzato, R. Hadsell, M.~F. Balcan, and H. Lin,
  editors, {\em Advances in Neural Information Processing Systems}, volume~33,
  pages 9912--9924. Curran Associates, Inc., 2020.

\bibitem{chen2020simple}
Ting Chen, Simon Kornblith, Mohammad Norouzi, and Geoffrey Hinton.
\newblock A simple framework for contrastive learning of visual
  representations.
\newblock In Hal~Daumé III and Aarti Singh, editors, {\em Proceedings of the
  37th International Conference on Machine Learning}, volume 119 of {\em
  Proceedings of Machine Learning Research}, pages 1597--1607. PMLR, 13--18 Jul
  2020.

\bibitem{quadtruplet}
Weihua Chen, Xiaotang Chen, Jianguo Zhang, and Kaiqi Huang.
\newblock Beyond triplet loss: a deep quadruplet network for person
  re-identification.
\newblock In {\em Proceedings of the IEEE Conference on Computer Vision and
  Pattern Recognition}, 2017.

\bibitem{metabaseline}
Yinbo Chen, Xiaolong Wang, Zhuang Liu, Huijuan Xu, and Trevor Darrell.
\newblock A new meta-baseline for few-shot learning.
\newblock {\em CoRR}, abs/2003.04390, 2020.

\bibitem{imagenet}
J. Deng, W. Dong, R. Socher, L.-J. Li, K. Li, and L. Fei-Fei.
\newblock {ImageNet: A Large-Scale Hierarchical Image Database}.
\newblock In {\em IEEE Conference on Computer Vision and Pattern Recognition
  (CVPR)}, 2009.

\bibitem{arcface}
Jiankang Deng, Jia Guo, Niannan Xue, and Stefanos Zafeiriou.
\newblock Arcface: Additive angular margin loss for deep face recognition.
\newblock In {\em Proceedings of the IEEE/CVF Conference on Computer Vision and
  Pattern Recognition (CVPR)}, June 2019.

\bibitem{bert}
Jacob Devlin, Ming-Wei Chang, Kenton Lee, and Kristina Toutanova.
\newblock {BERT}: Pre-training of deep bidirectional transformers for language
  understanding.
\newblock In {\em Proceedings of the 2019 Conference of the North {A}merican
  Chapter of the Association for Computational Linguistics: Human Language
  Technologies, Volume 1 (Long and Short Papers)}, pages 4171--4186,
  Minneapolis, Minnesota, June 2019. Association for Computational Linguistics.

\bibitem{vit}
Alexey Dosovitskiy, Lucas Beyer, Alexander Kolesnikov, Dirk Weissenborn,
  Xiaohua Zhai, Thomas Unterthiner, Mostafa Dehghani, Matthias Minderer, Georg
  Heigold, Sylvain Gelly, Jakob Uszkoreit, and Neil Houlsby.
\newblock An image is worth 16x16 words: Transformers for image recognition at
  scale.
\newblock In {\em International Conference on Learning Representations}, 2021.

\bibitem{daml}
Yueqi Duan, Wenzhao Zheng, Xudong Lin, Jiwen Lu, and Jie Zhou.
\newblock Deep adversarial metric learning.
\newblock In {\em The IEEE Conference on Computer Vision and Pattern
  Recognition (CVPR)}, June 2018.

\bibitem{ood_1}
Logan Engstrom, Brandon Tran, Dimitris Tsipras, Ludwig Schmidt, and Aleksander
  Madry.
\newblock Exploring the landscape of spatial robustness.
\newblock In Kamalika Chaudhuri and Ruslan Salakhutdinov, editors, {\em
  Proceedings of the 36th International Conference on Machine Learning},
  volume~97 of {\em Proceedings of Machine Learning Research}, pages
  1802--1811. PMLR, 09--15 Jun 2019.

\bibitem{fehervari2019unbiased}
Istvan Fehervari, Avinash Ravichandran, and Srikar Appalaraju.
\newblock Unbiased evaluation of deep metric learning algorithms, 2019.

\bibitem{maml}
Chelsea Finn, Pieter Abbeel, and Sergey Levine.
\newblock Model-agnostic meta-learning for fast adaptation of deep networks.
\newblock In Doina Precup and Yee~Whye Teh, editors, {\em Proceedings of the
  34th International Conference on Machine Learning}, volume~70 of {\em
  Proceedings of Machine Learning Research}, pages 1126--1135. PMLR, 06--11 Aug
  2017.

\bibitem{htl}
Weifeng Ge.
\newblock Deep metric learning with hierarchical triplet loss.
\newblock In {\em Proceedings of the European Conference on Computer Vision
  (ECCV)}, pages 269--285, 2018.

\bibitem{nca}
Jacob Goldberger, Geoffrey~E Hinton, Sam Roweis, and Russ~R Salakhutdinov.
\newblock Neighbourhood components analysis.
\newblock In L. Saul, Y. Weiss, and L. Bottou, editors, {\em Advances in Neural
  Information Processing Systems}, volume~17. MIT Press, 2005.

\bibitem{contrastive}
Raia Hadsell, Sumit Chopra, and Yann LeCun.
\newblock Dimensionality reduction by learning an invariant mapping.
\newblock In {\em Proceedings of the IEEE Conference on Computer Vision and
  Pattern Recognition}, 2006.

\bibitem{smartmining}
Ben Harwood, BG Kumar, Gustavo Carneiro, Ian Reid, Tom Drummond, et~al.
\newblock Smart mining for deep metric learning.
\newblock In {\em Proceedings of the IEEE International Conference on Computer
  Vision}, pages 2821--2829, 2017.

\bibitem{moco}
Kaiming He, Haoqi Fan, Yuxin Wu, Saining Xie, and Ross Girshick.
\newblock Momentum contrast for unsupervised visual representation learning.
\newblock In {\em Proceedings of the IEEE/CVF Conference on Computer Vision and
  Pattern Recognition (CVPR)}, June 2020.

\bibitem{resnet}
Kaiming He, Xiangyu Zhang, Shaoqing Ren, and Jian Sun.
\newblock Deep residual learning for image recognition.
\newblock In {\em Proceedings of the IEEE conference on computer vision and
  pattern recognition}, pages 770--778, 2016.

\bibitem{ood_2}
Dan Hendrycks and Thomas Dietterich.
\newblock Benchmarking neural network robustness to common corruptions and
  perturbations.
\newblock In {\em International Conference on Learning Representations}, 2019.

\bibitem{genlifted}
Alexander {Hermans}, Lucas {Beyer}, and Bastian {Leibe}.
\newblock {In Defense of the Triplet Loss for Person Re-Identification}.
\newblock {\em arXiv e-prints}, page arXiv:1703.07737, Mar. 2017.

\bibitem{fid}
Martin Heusel, Hubert Ramsauer, Thomas Unterthiner, Bernhard Nessler, and Sepp
  Hochreiter.
\newblock Gans trained by a two time-scale update rule converge to a local nash
  equilibrium, 2017.

\bibitem{horde}
Pierre Jacob, David Picard, Aymeric Histace, and Edouard Klein.
\newblock Metric learning with horde: High-order regularizer for deep
  embeddings.
\newblock In {\em The IEEE Conference on Computer Vision and Pattern
  Recognition (CVPR)}, 2019.

\bibitem{recall}
Herve Jegou, Matthijs Douze, and Cordelia Schmid.
\newblock Product quantization for nearest neighbor search.
\newblock {\em IEEE transactions on pattern analysis and machine intelligence},
  33(1):117--128, 2011.

\bibitem{kim2020proxy}
Sungyeon Kim, Dongwon Kim, Minsu Cho, and Suha Kwak.
\newblock Proxy anchor loss for deep metric learning.
\newblock In {\em Proceedings of the IEEE/CVF Conference on Computer Vision and
  Pattern Recognition (CVPR)}, June 2020.

\bibitem{abe}
Wonsik Kim, Bhavya Goyal, Kunal Chawla, Jungmin Lee, and Keunjoo Kwon.
\newblock Attention-based ensemble for deep metric learning.
\newblock In {\em Proceedings of the European Conference on Computer Vision
  (ECCV)}, 2018.

\bibitem{adam}
Diederik~P Kingma and Jimmy Ba.
\newblock Adam: A method for stochastic optimization.
\newblock 2015.

\bibitem{koh2021wilds}
Pang~Wei Koh, Shiori Sagawa, Henrik Marklund, Sang~Michael Xie, Marvin Zhang,
  Akshay Balsubramani, Weihua Hu, Michihiro Yasunaga, Richard~Lanas Phillips,
  Irena Gao, Tony Lee, Etienne David, Ian Stavness, Wei Guo, Berton~A.
  Earnshaw, Imran~S. Haque, Sara Beery, Jure Leskovec, Anshul Kundaje, Emma
  Pierson, Sergey Levine, Chelsea Finn, and Percy Liang.
\newblock Wilds: A benchmark of in-the-wild distribution shifts, 2021.

\bibitem{bit}
Alexander Kolesnikov, Lucas Beyer, Xiaohua Zhai, Joan Puigcerver, Jessica Yung,
  Sylvain Gelly, and Neil Houlsby.
\newblock Big transfer (bit): General visual representation learning.
\newblock In Andrea Vedaldi, Horst Bischof, Thomas Brox, and Jan-Michael Frahm,
  editors, {\em Computer Vision -- ECCV 2020}, pages 491--507, Cham, 2020.
  Springer International Publishing.

\bibitem{cars196}
Jonathan Krause, Michael Stark, Jia Deng, and Li Fei-Fei.
\newblock 3d object representations for fine-grained categorization.
\newblock In {\em Proceedings of the IEEE International Conference on Computer
  Vision Workshops}, pages 554--561, 2013.

\bibitem{ood_4}
David Krueger, Ethan Caballero, Joern-Henrik Jacobsen, Amy Zhang, Jonathan
  Binas, Dinghuai Zhang, Remi~Le Priol, and Aaron Courville.
\newblock Out-of-distribution generalization via risk extrapolation (rex),
  2021.

\bibitem{metaoptnet}
Kwonjoon Lee, Subhransu Maji, Avinash Ravichandran, and Stefano Soatto.
\newblock Meta-learning with differentiable convex optimization.
\newblock In {\em Proceedings of the IEEE/CVF Conference on Computer Vision and
  Pattern Recognition (CVPR)}, June 2019.

\bibitem{dvml}
Xudong Lin, Yueqi Duan, Qiyuan Dong, Jiwen Lu, and Jie Zhou.
\newblock Deep variational metric learning.
\newblock In {\em The European Conference on Computer Vision (ECCV)}, September
  2018.

\bibitem{sphereface}
Weiyang Liu, Yandong Wen, Zhiding Yu, Ming Li, Bhiksha Raj, and Le Song.
\newblock Sphereface: Deep hypersphere embedding for face recognition.
\newblock {\em IEEE Conference on Computer Vision and Pattern Recognition
  (CVPR)}, 2017.

\bibitem{roberta}
Yinhan Liu, Myle Ott, Naman Goyal, Jingfei Du, Mandar Joshi, Danqi Chen, Omer
  Levy, Mike Lewis, Luke Zettlemoyer, and Veselin Stoyanov.
\newblock Roberta: A robustly optimized bert pretraining approach, 2019.
\newblock cite arxiv:1907.11692.

\bibitem{torchvision}
S\'{e}bastien Marcel and Yann Rodriguez.
\newblock Torchvision the machine-vision package of torch.
\newblock In {\em Proceedings of the 18th ACM International Conference on
  Multimedia}, MM '10, page 1485–1488, New York, NY, USA, 2010. Association
  for Computing Machinery.

\bibitem{umap}
Leland McInnes, John Healy, Nathaniel Saul, and Lukas Grossberger.
\newblock Umap: Uniform manifold approximation and projection.
\newblock {\em The Journal of Open Source Software}, 3(29):861, 2018.

\bibitem{milbich2017unsupervised}
Timo Milbich, Miguel Bautista, Ekaterina Sutter, and Bj{\"o}rn Ommer.
\newblock Unsupervised video understanding by reconciliation of posture
  similarities.
\newblock In {\em Proceedings of the International Conference on Computer
  Vision (ICCV)}, 2017.

\bibitem{milbich2020reliablerelations}
Timo Milbich, Omair Ghori, and Björn Ommer.
\newblock Unsupervised representation learning by discovering reliable image
  relations.
\newblock {\em Pattern Recognition}, 102, June 2020.

\bibitem{milbich2020diva}
Timo Milbich, Karsten Roth, Homanga Bharadhwaj, Samarth Sinha, Yoshua Bengio,
  Bj{\"o}rn Ommer, and Joseph~Paul Cohen.
\newblock Diva: Diverse visual feature aggregation for deep metric learning.
\newblock In Andrea Vedaldi, Horst Bischof, Thomas Brox, and Jan-Michael Frahm,
  editors, {\em Computer Vision -- ECCV 2020}, pages 590--607, Cham, 2020.
  Springer International Publishing.

\bibitem{milbich2020sharing}
Timo {Milbich}, Karsten {Roth}, Biagio {Brattoli}, and Bj{\"o}rn {Ommer}.
\newblock Sharing matters for generalization in deep metric learning.
\newblock {\em IEEE Transactions on Pattern Analysis and Machine Intelligence},
  2020.

\bibitem{pretextmisra}
Ishan Misra and Laurens van~der Maaten.
\newblock Self-supervised learning of pretext-invariant representations.
\newblock In {\em Proceedings of the IEEE/CVF Conference on Computer Vision and
  Pattern Recognition (CVPR)}, June 2020.

\bibitem{proxynca}
Yair Movshovitz-Attias, Alexander Toshev, Thomas~K Leung, Sergey Ioffe, and
  Saurabh Singh.
\newblock No fuss distance metric learning using proxies.
\newblock In {\em Proceedings of the IEEE International Conference on Computer
  Vision}, pages 360--368, 2017.

\bibitem{musgrave2020metric}
Kevin Musgrave, Serge Belongie, and Ser-Nam Lim.
\newblock A metric learning reality check, 2020.

\bibitem{lifted}
Hyun Oh~Song, Yu Xiang, Stefanie Jegelka, and Silvio Savarese.
\newblock Deep metric learning via lifted structured feature embedding.
\newblock In {\em Proceedings of the IEEE Conference on Computer Vision and
  Pattern Recognition}, pages 4004--4012, 2016.

\bibitem{bier}
Michael Opitz, Georg Waltner, Horst Possegger, and Horst Bischof.
\newblock Bier-boosting independent embeddings robustly.
\newblock In {\em Proceedings of the IEEE International Conference on Computer
  Vision}, pages 5189--5198, 2017.

\bibitem{abier}
Michael Opitz, Georg Waltner, Horst Possegger, and Horst Bischof.
\newblock Deep metric learning with bier: Boosting independent embeddings
  robustly.
\newblock {\em IEEE transactions on pattern analysis and machine intelligence},
  2018.

\bibitem{pytorch}
Adam Paszke, Sam Gross, Soumith Chintala, Gregory Chanan, Edward Yang, Zachary
  DeVito, Zeming Lin, Alban Desmaison, Luca Antiga, and Adam Lerer.
\newblock Automatic differentiation in pytorch.
\newblock In {\em NIPS-W}, 2017.

\bibitem{softriple}
Qi Qian, Lei Shang, Baigui Sun, Juhua Hu, Hao Li, and Rong Jin.
\newblock Softtriple loss: Deep metric learning without triplet sampling.
\newblock In {\em Proceedings of the IEEE/CVF International Conference on
  Computer Vision (ICCV)}, October 2019.

\bibitem{clip}
Alec Radford, Jong~Wook Kim, Chris Hallacy, Aditya Ramesh, Gabriel Goh,
  Sandhini Agarwal, Girish Sastry, Amanda Askell, Pamela Mishkin, Jack Clark,
  Gretchen Krueger, and Ilya Sutskever.
\newblock Learning transferable visual models from natural language
  supervision.
\newblock {\em CoRR}, abs/2103.00020, 2021.

\bibitem{anil}
Aniruddh Raghu, Maithra Raghu, Samy Bengio, and Oriol Vinyals.
\newblock Rapid learning or feature reuse? towards understanding the
  effectiveness of maml.
\newblock In {\em International Conference on Learning Representations}, 2020.

\bibitem{ood_3}
Benjamin Recht, Rebecca Roelofs, Ludwig Schmidt, and Vaishaal Shankar.
\newblock Do {I}mage{N}et classifiers generalize to {I}mage{N}et?
\newblock In Kamalika Chaudhuri and Ruslan Salakhutdinov, editors, {\em
  Proceedings of the 36th International Conference on Machine Learning},
  volume~97 of {\em Proceedings of Machine Learning Research}, pages
  5389--5400. PMLR, 09--15 Jun 2019.

\bibitem{im21k_check}
Tal Ridnik, Emanuel~Ben Baruch, Asaf Noy, and Lihi Zelnik{-}Manor.
\newblock Imagenet-21k pretraining for the masses.
\newblock {\em CoRR}, abs/2104.10972, 2021.

\bibitem{mic}
Karsten Roth, Biagio Brattoli, and Bjorn Ommer.
\newblock Mic: Mining interclass characteristics for improved metric learning.
\newblock In {\em Proceedings of the IEEE International Conference on Computer
  Vision}, pages 8000--8009, 2019.

\bibitem{roth2020pads}
Karsten Roth, Timo Milbich, and Bjorn Ommer.
\newblock Pads: Policy-adapted sampling for visual similarity learning.
\newblock In {\em Proceedings of the IEEE/CVF Conference on Computer Vision and
  Pattern Recognition (CVPR)}, June 2020.

\bibitem{roth2020s2sd}
Karsten Roth, Timo Milbich, Bj{\"{o}}rn Ommer, Joseph~Paul Cohen, and Marzyeh
  Ghassemi.
\newblock {S2SD:} simultaneous similarity-based self-distillation for deep
  metric learning.
\newblock {\em CoRR}, abs/2009.08348, 2020.

\bibitem{roth2020revisiting}
Karsten Roth, Timo Milbich, Samarth Sinha, Prateek Gupta, Bjorn Ommer, and
  Joseph~Paul Cohen.
\newblock Revisiting training strategies and generalization performance in deep
  metric learning.
\newblock In Hal~Daumé III and Aarti Singh, editors, {\em Proceedings of the
  37th International Conference on Machine Learning}, volume 119 of {\em
  Proceedings of Machine Learning Research}, pages 8242--8252. PMLR, 13--18 Jul
  2020.

\bibitem{ood_5}
Karsten Roth, Latha Pemula, Joaquin Zepeda, Bernhard Sch{\"{o}}lkopf, Thomas
  Brox, and Peter~V. Gehler.
\newblock Towards total recall in industrial anomaly detection.
\newblock {\em CoRR}, abs/2106.08265, 2021.

\bibitem{Sanakoyeu_2019_CVPR}
Artsiom Sanakoyeu, Vadim Tschernezki, Uta Buchler, and Bjorn Ommer.
\newblock Divide and conquer the embedding space for metric learning.
\newblock In {\em The IEEE Conference on Computer Vision and Pattern
  Recognition (CVPR)}, 2019.

\bibitem{semihard}
Florian Schroff, Dmitry Kalenichenko, and James Philbin.
\newblock Facenet: A unified embedding for face recognition and clustering.
\newblock In {\em Proceedings of the IEEE conference on computer vision and
  pattern recognition}, pages 815--823, 2015.

\bibitem{intrabatch}
Jenny Seidenschwarz, Ismail Elezi, and Laura Leal-Taixé.
\newblock Learning intra-batch connections for deep metric learning, 2021.

\bibitem{sinha2020uniform}
Samarth Sinha, Karsten Roth, Anirudh Goyal, Marzyeh Ghassemi, Hugo Larochelle,
  and Animesh Garg.
\newblock Uniform priors for data-efficient transfer, 2020.

\bibitem{protonet}
Jake Snell, Kevin Swersky, and Richard Zemel.
\newblock Prototypical networks for few-shot learning.
\newblock In I. Guyon, U.~V. Luxburg, S. Bengio, H. Wallach, R. Fergus, S.
  Vishwanathan, and R. Garnett, editors, {\em Advances in Neural Information
  Processing Systems}, volume~30. Curran Associates, Inc., 2017.

\bibitem{npairs}
Kihyuk Sohn.
\newblock Improved deep metric learning with multi-class n-pair loss objective.
\newblock In {\em Advances in Neural Information Processing Systems}, pages
  1857--1865, 2016.

\bibitem{surez2018tutorial}
Juan~Luis Suárez, Salvador García, and Francisco Herrera.
\newblock A tutorial on distance metric learning: Mathematical foundations,
  algorithms and experiments, 2018.

\bibitem{efficientnet}
Mingxing Tan and Quoc Le.
\newblock {E}fficient{N}et: Rethinking model scaling for convolutional neural
  networks.
\newblock In Kamalika Chaudhuri and Ruslan Salakhutdinov, editors, {\em
  Proceedings of the 36th International Conference on Machine Learning},
  volume~97 of {\em Proceedings of Machine Learning Research}, pages
  6105--6114. PMLR, 09--15 Jun 2019.

\bibitem{proxypp}
Eu~Wern Teh, Terrance DeVries, and Graham~W. Taylor.
\newblock Proxynca++: Revisiting and revitalizing proxy neighborhood component
  analysis.
\newblock In Andrea Vedaldi, Horst Bischof, Thomas Brox, and Jan-Michael Frahm,
  editors, {\em Computer Vision -- ECCV 2020}, pages 448--464, Cham, 2020.
  Springer International Publishing.

\bibitem{tian2019contrastive}
Yonglong Tian, Dilip Krishnan, and Phillip Isola.
\newblock Contrastive representation distillation.
\newblock In {\em International Conference on Learning Representations}, 2020.

\bibitem{tian2020rethinking}
Yonglong Tian, Yue Wang, Dilip Krishnan, Joshua~B. Tenenbaum, and Phillip
  Isola.
\newblock Rethinking few-shot image classification: A good embedding is all you
  need?
\newblock In Andrea Vedaldi, Horst Bischof, Thomas Brox, and Jan-Michael Frahm,
  editors, {\em Computer Vision -- ECCV 2020}, pages 266--282, Cham, 2020.
  Springer International Publishing.

\bibitem{fewshot_info}
Eleni Triantafillou, Richard Zemel, and Raquel Urtasun.
\newblock Few-shot learning through an information retrieval lens.
\newblock In I. Guyon, U.~V. Luxburg, S. Bengio, H. Wallach, R. Fergus, S.
  Vishwanathan, and R. Garnett, editors, {\em Advances in Neural Information
  Processing Systems}, volume~30. Curran Associates, Inc., 2017.

\bibitem{cub200-2011}
C. Wah, S. Branson, P. Welinder, P. Perona, and S. Belongie.
\newblock {The Caltech-UCSD Birds-200-2011 Dataset}.
\newblock Technical Report CNS-TR-2011-001, California Institute of Technology,
  2011.

\bibitem{angular}
Jian Wang, Feng Zhou, Shilei Wen, Xiao Liu, and Yuanqing Lin.
\newblock Deep metric learning with angular loss.
\newblock In {\em Proceedings of the IEEE International Conference on Computer
  Vision}, pages 2593--2601, 2017.

\bibitem{multisimilarity}
Xun Wang, Xintong Han, Weilin Huang, Dengke Dong, and Matthew~R. Scott.
\newblock Multi-similarity loss with general pair weighting for deep metric
  learning.
\newblock In {\em Proceedings of the IEEE/CVF Conference on Computer Vision and
  Pattern Recognition (CVPR)}, June 2019.

\bibitem{rw2019timm}
Ross Wightman.
\newblock Pytorch image models.
\newblock \url{https://github.com/rwightman/pytorch-image-models}, 2019.

\bibitem{margin}
Chao-Yuan Wu, R Manmatha, Alexander~J Smola, and Philipp Krahenbuhl.
\newblock Sampling matters in deep embedding learning.
\newblock In {\em Proceedings of the IEEE International Conference on Computer
  Vision}, pages 2840--2848, 2017.

\bibitem{epshn}
Hong Xuan, Abby Stylianou, and Robert Pless.
\newblock Improved embeddings with easy positive triplet mining.
\newblock In {\em Proceedings of the IEEE/CVF Winter Conference on Applications
  of Computer Vision (WACV)}, March 2020.

\bibitem{zhai2018classification}
Andrew Zhai and Hao-Yu Wu.
\newblock Classification is a strong baseline for deep metric learning, 2018.

\bibitem{hardness-aware}
Wenzhao Zheng, Zhaodong Chen, Jiwen Lu, and Jie Zhou.
\newblock Hardness-aware deep metric learning.
\newblock {\em The IEEE Conference on Computer Vision and Pattern Recognition
  (CVPR)}, 2019.

\end{thebibliography}
}

%%%%%%%%%%%%%%%%%%%%%%%%%%%%%%%%%%%%%%%%%%%%%%%%%%%%%%%%%%%%
\newpage
\appendix
\title{Supplementary: Characterizing Generalization under Out-Of-Distribution Shifts in Deep Metric Learning}
\author{}
\maketitlepage
\setcounter{equation}{0}
\setcounter{figure}{0}
\setcounter{table}{0}
\setcounter{page}{1}
\makeatletter
\renewcommand{\theequation}{S\arabic{equation}}
\renewcommand{\thefigure}{S\arabic{figure}}
\renewcommand{\thetable}{S\arabic{table}}

%%%%%%%%%%%%%%%%%%%%%%%%%%%%
\section{Analyzing the model bias for selecting train-test splits}
\begin{figure}[h]
    \centering
    \includegraphics[width=1\textwidth]{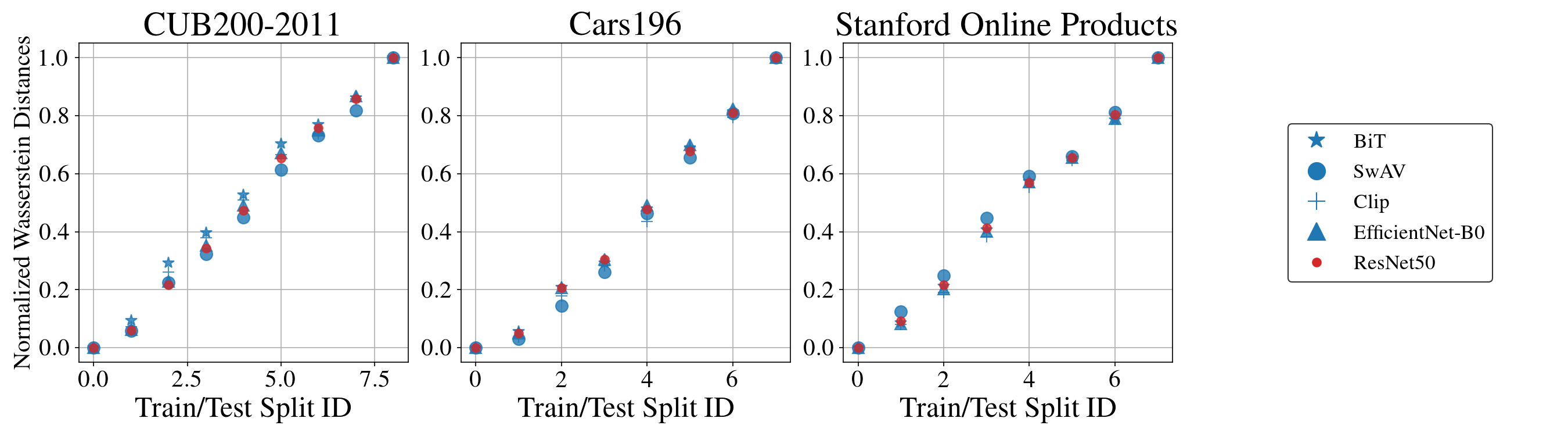}
    \caption{\textit{Normalized FID progession for ooDML train-test splits using different training models and networks.} Values are normalized for comparability of FID progression, as FID scores are not upper bounded and as such, absolute values for different networks and pretraining methods differ.}
    \label{supp_fig:fid_progression_comp}
\end{figure}
\begin{figure}[h]
    \centering
    \includegraphics[width=0.9\textwidth]{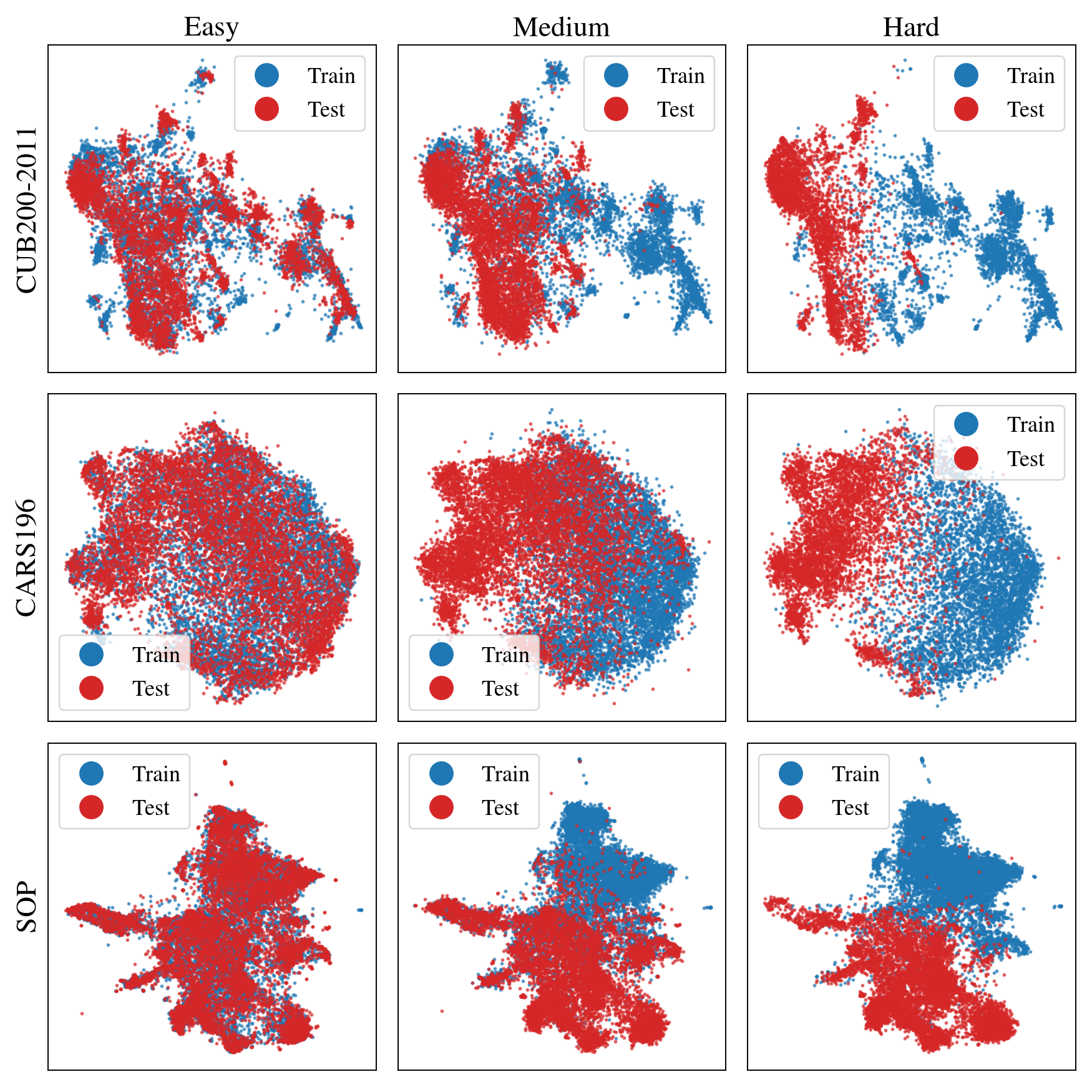}
    \caption{\textit{UMAPs} for easy (split-id 1), medium (split-id 5) and hard (split-id 8/9) for all benchmarks using a ResNet50 backbone pretrained on ImageNet. As can be seen, our iterative class swapping (and removal) procedure (cf. Sec. 3.2 main paper) creates splits in which training and test distributions become increasingly disjoint. Note that while we shifted full classes for semantic consistency, each point corresponds to a single sample (for SOP, random subsampling to 20000 total points was performed).}
    \label{supp_fig:umap_progression}
\end{figure}
To analyze the impact of the network architecture, pretraining method and training data, respectively the learned feature representations, on the construction of train-test splits and the entailed difficulties, we repeat our class swapping and removal procedure introduced in Section \ref{sec:gen_study} in the main paper using different self-supervised models. Subsequently, we select train-test splits from the same iteration steps. Fig.~\ref{supp_fig:fid_progression_comp} compares the progression of distribution shifts based on FID scores normalized to the $[0,1]$ interval for valid comparison. We observe that across all pretrained models, the general FID progressions and sampled train-test splits exhibit very similar learning problem difficulties, indicating that our sampling procedure is robust to the choice of readily available, state-of-the art self-supervised pretrained models.

%%%%%%%%%%%%%%%%%%%%%%%%%%%%
\section{Further Details regarding the Experimental Setup}

\paragraph{Datasets.}
In total, we utilized three widely used Deep Metric Learning benchmarks: (1) CUB200-2011 \cite{cub200-2011}, which comprises a total of 11,788 images over 200 classes of birds, (2) CARS196 \cite{cars196} containing 16,185 images of cars distributed over 196 classes and (3) Stanford Online Products (SOP) \cite{lifted}, which introduced 120,053 product images over 22,634 total classes. For CUB200-2011 and CARS196, default splits are simply generated by selecting the last half of the alphabetically sorted classes as test samples, whereas SOP provides a predefined split with 11318 training and 11316 test classes.

\paragraph{Training details.}
For our implementation, we leveraged the PyTorch framework \cite{pytorch}. For training, in all cases, training images were randomly resized and cropped to $224\times 224$, whereas for testing images were resized to $256\times 256$ and center cropped to $224\times 224$. Optimization was performed with Adam \cite{adam} with learning rate of $10^{-5}$ and weight decay of $3\cdot 10^{-4}$. Batchsizes where chosen within the range of [86, 112] depending on the size of the utilized backbone network. For default DML ResNet-architectures, we follow previous literature \cite{margin,kim2020proxy,roth2020revisiting,musgrave2020metric} and freeze Batch-Normalization layers during training. We consistently use an embedding dimensionality of $512$ for comparability. For DiVA \cite{milbich2020diva}, S2SD \cite{roth2020s2sd} and ProxyAnchor \cite{kim2020proxy}, parameter choices were set to default values given in the original publications, with small grid searches done to allow for adaptation to backbone changes. For all other remaining objectives, parameter choices were adapted from \cite{roth2020revisiting}, who provide a hyperparameter selection for best comparability of methods. All experiments were performed on GPU servers containing NVIDIA P100, T4 and Titan X, with results always averaged over multiple seeds - in the case of our objective study five random seeds were utilized, whereas for other ablation-type studies at least three seeds were utilized. These settings are used throughout our study. For the few-shot experiments, the same pipeline parameters were utilized with changes noted in the respective section.
\\
Pretrained network weights for ResNet-architectures where taken from \textsf{torchvision} \cite{torchvision}, EfficientNet and BiT weights from \textsf{timm} \cite{rw2019timm} and SwAV and CLIP pretrained weights from the respective official repositories (\cite{swav} and \cite{clip}).

\paragraph{FID scores between \textit{ooDML }data splits.}
In Tab.~\ref{supp_tab:fid_oodml} we show the measured FID scores between each train-test split of our \textit{ooDML} for the  CUB200-2011, CARS196 and SOP benchmarks, respectively.

\begin{table}[t]
    \caption{FID scores between train-test splits in our ooDML benchmark. For details on creating train-test splits constituting the \textit{ooDML} benchmark, please see Sec. \ref{sec:gen_study} in main paper.}
    \centering
    \resizebox{0.8\textwidth}{!}{   
    \begin{tabular}{l||c|c|c|c|c|c|c|c|c}
         \toprule
         \textbf{Dataset $\downarrow$ split-ID $\rightarrow$} & 1 & 2 & 3 & 4 & 5 & 6 & 7 & 8 & 9 \\
         \midrule
         CUB200-2011 & 19.2 & 28.5 & 52.6 & 72.2 & 92.5 & 120.4 & 136.5 & 152.0 & 173.9 \\
         CARS196 & 8.6 & 14.3 & 32.2 & 43.6 & 63.3 & 86.5 & 101.2 & 123.0 & - \\
         SOP & 3.4 & 24.6 & 53.5 & 99.4 & 135.5 & 155.3 & 189.8 & 235.1 & -\\
         \bottomrule
    \end{tabular}}
\label{supp_tab:fid_oodml}    
\end{table}

\paragraph{Qualitative introspection of \textit{ooDML} train-test splits using UMAP.} 
Fig.~\ref{supp_fig:umap_progression} visualizes the distribution shift between train-test splits from our proposed \textit{ooDML} benchmark using the UMAP~\cite{umap} algorithm. For each dataset we show examples for an easy, medium and hard train-test split. Indeed, the distributional shift train to test data is increasing consistently, as indicated by our monotonically increasing FID progressions.

%This section includes UMAP progressions for each splits as well as class split examples (using e.g. the superclass context on SOP).

\section{On the limits of OOD generalization in Deep Metric Learning}
\begin{figure*}[t]
    \centering
    \includegraphics[width=1\textwidth]{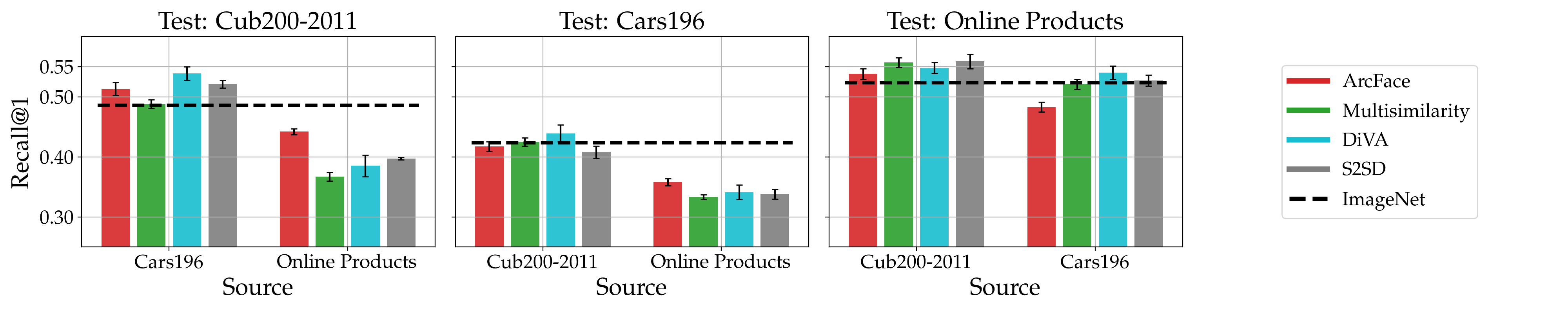}
    \caption{\textit{Out-of-Domain Generalization.} Each plot showcases transfer performance from the training dataset (\textit{source}) to a test dataset from a novel domain (\textit{test}). The dashed line represents baseline performance achieved by ResNet50 pretrained on ImageNet.}
    \label{supp_fig:new_domain_transfer}
\end{figure*}
\begin{table*}[t]
    \caption{FID scores between training and test sets across different datasets compared to the highest FID measured by our generated train-test splits.}
    \label{tab:fid_diff_datasets}
    
    \centering
    \resizebox{1\textwidth}{!}{   
    \begin{tabular}{l||c|c|c|c|c|c||c}
            \toprule
         \textbf{Direction (train$\rightarrow$test)} & CUB$\rightarrow$CARS & CARS$\rightarrow$CUB & CUB$\rightarrow$SOP & SOP$\rightarrow$CUB & CARS$\rightarrow$SOP & SOP$\rightarrow$CARS & Max. \textit{ooDML}\\
         \hline
         \textbf{FID} & 349 & 359 & 359 & 370 & 386 & 376 & 155 (235) \\
         \bottomrule
    \end{tabular}}
\end{table*}
To investigate how well representations $\phi$ learned by DML approaches transfer \textit{across} benchmark datasets, we train our models on the default training dataset of one benchmark and evaluate them on the default test set of another. Tab.~\ref{tab:fid_diff_datasets} first illustrates the FID scores for all pairwise combinations using the CUB200-2011, CARS196 and SOP datasets. We find all FID scores exceed the previously considered learning problems in our \textit{ooDML} benchmark by far. 
However, the fact that FID scores are relatively close to another despite large semantic differences between datasets may indicate that FID based on our utilised FID estimator (Sec. \ref{sec:gen_study}) may have reached its limit as a distributional shift indicator, thus not being sufficiently sensitive. Fig.~\ref{supp_fig:new_domain_transfer} summarizes the generalization performances for different DML approaches on this experimental setting. 
As can be seen, there are only a few cases where $\phi$ offers a benefit over the ResNet50 ImageNet baseline, indicating that generalization of DML approaches is primarily limited to shifts within a data domains. 
Beyond these limits, generic representations learned by self-supervised learning may offer better zero-shot generalization, as also discussed on Sec. 4.4.

\section{Additional Experimental Results}

\subsection{Zero-shot generalization under varying distribution shifts}
\begin{figure*}[t]
    \centering
    \includegraphics[width=0.95\textwidth]{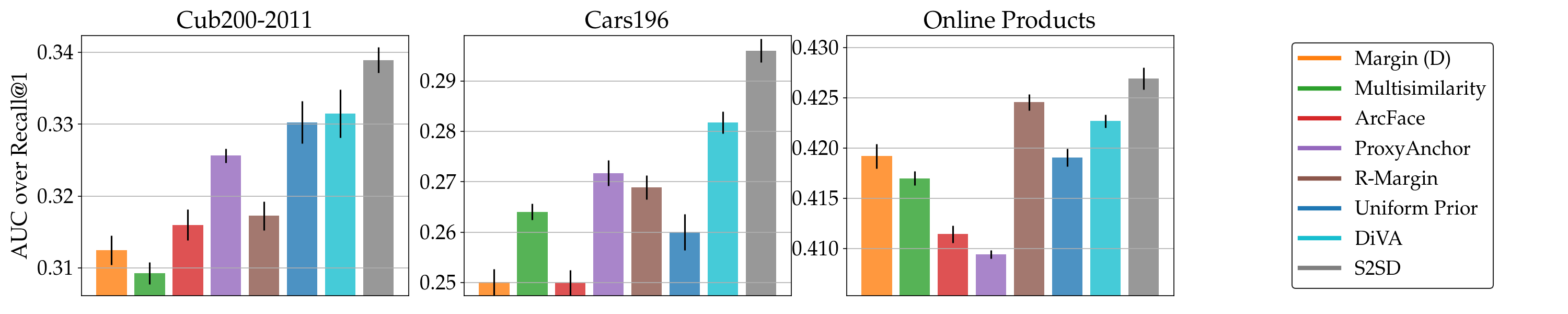}
    \caption{\textit{Comparison of DML methods via AGS based on mAP@1000 across benchmarks.} To compute AGS (cf. Sec. 4.2 main paper), we aggregate the mAP@1000 performances in Tab.~\ref{supp_tab:baseline_prog_full_cub}-\ref{supp_tab:baseline_prog_full_sop} across all train-test distribution shifts of our proposed \textit{ooDML} benchmark using the Area-Under-the-Curve metric.}
    \label{supp_fig:auc_shift_comparison}
\end{figure*}

This section provides additional results for the experiments presented in Sec. \label{subsec:base_exp_results}. To this end, we provide the exact performance values used to visualize Fig. \ref{fig:total_shift_comparison} in the main paper in Tab.~\ref{supp_tab:baseline_prog_full_cub}-\ref{supp_tab:baseline_prog_full_sop}. 
For the comparison based on the Aggregated Generalization Score (AGS) introduced in Sec. \ref{subsec:base_exp_results} in the main paper, Tab.~\ref{tab:auc_comparison} provides the empirical results both for AGS computed based on Recall@1 and mAP@1000. For the latter, Fig.~\ref{supp_fig:auc_shift_comparison} summarizes AGS results using a bar plot similar to Fig. \ref{fig:auc_shift_comparison} in the main paper.

\begin{table*}[b]
    \caption{Results for Aggregated Generalization Score (AGS) (cf. Sec. 4.2 main paper) based on Recall@1 and mAP@1000 computed on the \textit{ooDML} benchmark. We show results for various DML methods averaged over over multiple runs.}
  \centering
\resizebox{0.75\textwidth}{!}{
 \footnotesize
  \setlength\tabcolsep{1.4pt}
  %\begin{tabular}{l|c|ccc|c}
  \begin{tabular}{l|c|c||c|c||c|c}
     \toprule
     \multicolumn{1}{l}{\textbf{Benchmark}$\rightarrow$} & \multicolumn{2}{c}{\textsc{CUB200-2011}} & \multicolumn{2}{c}{\textsc{CARS196}} & \multicolumn{2}{c}{\textsc{SOP}} \\
     \midrule
     \textbf{Approaches} $\downarrow$ \textbf{AUC} $\rightarrow$ & R@1 & mAP@1000 & R@1 & mAP@1000 & R@1 & mAP@1000\\
    \midrule
    \rowcolor{vvlightgray}
    Margin (D) \cite{margin} & $63.6\pm0.3$ & $31.2\pm0.2$ & $74.5\pm0.4$ & $25.0\pm0.3$ & $74.6\pm0.1$ & $41.9\pm0.1$\\
    Multisimilarity \cite{multisimilarity} & $64.3\pm0.3$ & $30.9\pm0.2$ & $76.1\pm0.2$ & $26.4\pm0.2$ & $74.6\pm0.1$ & $41.7\pm0.1$\\
    \rowcolor{vvlightgray}    
    ArcFace \cite{arcface} & $63.3\pm0.4$ & $31.6\pm0.2$ & $73.2\pm0.3$ & $25.0\pm0.3$ & $73.9\pm0.1$ & $41.1\pm0.1$\\
    ProxyAnchor \cite{kim2020proxy} & $65.1\pm0.2$ & $32.6\pm0.1$ & $76.6\pm0.2$ & $27.2\pm0.3$ & $74.0\pm0.1$ & $40.9\pm0.1$\\
    \rowcolor{vvlightgray}    
    R-Margin \cite{roth2020revisiting} & $65.4\pm0.3$ & $31.7\pm0.2$ & $77.3\pm0.3$ & $26.9\pm0.2$ & $74.9\pm0.1$ & $42.5\pm0.1$\\
    Uniform Prior & $65.7\pm0.5$ & $33.0\pm0.3$ & $75.8\pm0.4$ & $26.0\pm0.4$ & $74.6\pm0.1$ & $41.9\pm0.1$\\
    \rowcolor{vvlightgray}    
    DiVA \cite{milbich2020diva} & $66.4\pm0.3$ & $33.1\pm0.3$ & $78.6\pm0.3$ & $27.9\pm0.2$ & $75.0\pm0.1$ & $42.3\pm0.1$\\
    S2SD & $\mathbf{67.7}\pm0.3$ & $\mathbf{33.9}\pm0.2$ & $\mathbf{80.2}\pm0.2$ & $\mathbf{29.6}\pm0.2$ & $\mathbf{75.1}\pm0.1$ & $\mathbf{42.7}\pm0.1$\\
    \bottomrule
    \end{tabular}}
    \label{tab:auc_comparison}
\end{table*}

\begin{table*}[t]
    \caption{DML generalization performance measured by Recall@1 and mAP@1000 on each train-test split of our \textit{ooDML} benchmark for the CUB200-2011 dataset.}
  \centering
\resizebox{1\textwidth}{!}{
 \footnotesize
  \setlength\tabcolsep{1.4pt}
  %\begin{tabular}{l|c|ccc|c}
  \begin{tabular}{l|l||c|c|c|c|c|c|c|c|c}
     \toprule
     \textbf{Metric}  & \textbf{Method}$\downarrow$ | \textbf{Split (FID)}$\rightarrow$ & 1 (19.2) & 2 (28.5) & 3 (52.6) & 4 (72.2) & 5 (92.5) & 6 (120.4) & 7 (136.5) & 8 (152.0) & 9 (173.9)\\
    \midrule
    \multirow{8}{*}{R@1} & Margin (D) &  $76.20 \pm 0.13$ & $71.79 \pm 0.09$ & $65.78 \pm 0.05$ & $65.38 \pm 0.22$ & $63.30 \pm 0.58$ & $61.53 \pm 0.43$ & $59.95 \pm 0.27$ & $57.67 \pm 0.53$ & $58.59 \pm 0.27$ \\
    & Multisimilarity &  $76.44 \pm 0.42$ & $72.34 \pm 0.11$ & $66.36 \pm 0.24$ & $66.21 \pm 0.26$ & $64.18 \pm 0.40$ & $62.68 \pm 0.34$ & $60.77 \pm 0.23$ & $58.35 \pm 0.06$ & $58.20 \pm 0.51$ \\
    & ArcFace &  $76.10 \pm 0.28$ & $71.22 \pm 0.08$ & $65.19 \pm 0.38$ & $64.41 \pm 0.39$ & $62.53 \pm 0.05$ & $61.29 \pm 0.27$ & $60.28 \pm 0.66$ & $58.71 \pm 0.74$ & $58.53 \pm 0.41$ \\
    & ProxyAnchor &  $77.30 \pm 0.14$ & $72.95 \pm 0.13$ & $66.64 \pm 0.09$ & $66.39 \pm 0.13$ & $64.64 \pm 0.26$ & $63.03 \pm 0.07$ & $62.27 \pm 0.08$ & $60.25 \pm 0.25$ & $60.44 \pm 0.21$ \\
    & R-Margin (D) &  $77.04 \pm 0.72$ & $72.82 \pm 0.18$ & $67.10 \pm 0.28$ & $67.31 \pm 0.17$ & $65.31 \pm 0.20$ & $63.57 \pm 0.28$ & $62.14 \pm 0.27$ & $59.90 \pm 0.39$ & $60.52 \pm 0.42$ \\
    & Uniform Prior &  $76.53 \pm 0.30$ & $72.52 \pm 0.17$ & $67.67 \pm 0.36$ & $67.47 \pm 0.39$ & $65.42 \pm 0.53$ & $64.64 \pm 0.56$ & $62.76 \pm 0.31$ & $60.03 \pm 0.55$ & $60.84 \pm 0.32$ \\
    & S2SD &  $78.93 \pm 0.20$ & $75.20 \pm 0.33$ & $69.24 \pm 0.51$ & $68.70 \pm 0.26$ & $67.28 \pm 0.17$ & $66.16 \pm 0.43$ & $64.64 \pm 0.08$ & $62.93 \pm 0.20$ & $63.02 \pm 0.32$ \\
    & DiVA &  $77.74 \pm 0.25$ & $73.15 \pm 0.26$ & $67.97 \pm 0.26$ & $67.74 \pm 0.19$ & $66.04 \pm 0.36$ & $64.61 \pm 0.12$ & $63.14 \pm 0.57$ & $61.83 \pm 0.57$ & $61.91 \pm 0.36$ \\

    \cmidrule[1pt]{1-11}                

    \multirow{8}{*}{mAP@1000} & Margin (D) &  $44.76 \pm 0.27$ & $39.69 \pm 0.20$ & $34.31 \pm 0.09$ & $33.59 \pm 0.13$ & $30.64 \pm 0.29$ & $28.17 \pm 0.16$ & $26.72 \pm 0.31$ & $25.45 \pm 0.18$ & $26.18 \pm 0.33$ \\
    & Multisimilarity &  $44.21 \pm 0.11$ & $39.03 \pm 0.07$ & $33.79 \pm 0.01$ & $33.38 \pm 0.26$ & $30.75 \pm 0.19$ & $27.83 \pm 0.15$ & $26.58 \pm 0.15$ & $25.25 \pm 0.22$ & $25.29 \pm 0.23$ \\
    & ArcFace &  $45.45 \pm 0.23$ & $40.39 \pm 0.37$ & $34.39 \pm 0.14$ & $34.08 \pm 0.06$ & $31.06 \pm 0.15$ & $27.58 \pm 0.15$ & $27.31 \pm 0.44$ & $26.19 \pm 0.28$ & $26.90 \pm 0.22$ \\
    & ProxyAnchor &  $46.20 \pm 0.11$ & $40.99 \pm 0.14$ & $35.23 \pm 0.03$ & $34.89 \pm 0.13$ & $32.06 \pm 0.04$ & $28.88 \pm 0.06$ & $28.15 \pm 0.18$ & $27.31 \pm 0.18$ & $28.18 \pm 0.02$ \\
    & R-Margin (D) &  $44.76 \pm 0.41$ & $39.77 \pm 0.15$ & $34.28 \pm 0.06$ & $34.56 \pm 0.29$ & $31.46 \pm 0.11$ & $28.25 \pm 0.06$ & $27.24 \pm 0.25$ & $26.48 \pm 0.37$ & $26.64 \pm 0.39$ \\
    & Uniform Prior &  $46.00 \pm 0.62$ & $41.04 \pm 0.40$ & $36.07 \pm 0.23$ & $35.51 \pm 0.33$ & $32.57 \pm 0.18$ & $30.15 \pm 0.33$ & $28.64 \pm 0.20$ & $27.21 \pm 0.23$ & $27.73 \pm 0.46$ \\
    & S2SD &  $47.19 \pm 0.08$ & $42.27 \pm 0.40$ & $36.49 \pm 0.02$ & $36.22 \pm 0.19$ & $33.79 \pm 0.10$ & $30.69 \pm 0.08$ & $29.11 \pm 0.09$ & $28.51 \pm 0.42$ & $28.83 \pm 0.30$ \\
    & DiVA &  $46.74 \pm 0.51$ & $41.07 \pm 0.71$ & $35.80 \pm 0.24$ & $35.63 \pm 0.24$ & $32.63 \pm 0.20$ & $30.02 \pm 0.25$ & $28.33 \pm 0.31$ & $27.88 \pm 0.49$ & $28.86 \pm 0.30$ \\

    \bottomrule
    \end{tabular}}
    \label{supp_tab:baseline_prog_full_cub}
\end{table*}

\begin{table*}[t]
    \caption{DML generalization performance measured by Recall@1 and mAP@1000 on each train-test split of our \textit{ooDML} benchmark for the CARS196 dataset.}
  \centering
\resizebox{1\textwidth}{!}{
 \footnotesize
  \setlength\tabcolsep{1.4pt}
  %\begin{tabular}{l|c|ccc|c}
  \begin{tabular}{l|l||c|c|c|c|c|c|c|c}
     \toprule
     \textbf{Metric}  & \textbf{Method}$\downarrow$ | \textbf{Split (FID)}$\rightarrow$ & 1 (8.6) & 2 (14.3) & 3 (32.2) & 4 (43.6) & 5 (63.3) & 6 (86.5) & 7 (101.2) & 8 (123.0) \\
    \midrule
    \multirow{8}{*}{R@1} & Margin (D) &  $83.89 \pm 0.24$ & $82.99 \pm 0.15$ & $81.27 \pm 0.26$ & $78.95 \pm 0.20$ & $75.59 \pm 0.32$ & $69.97 \pm 0.61$ & $67.41 \pm 0.38$ & $64.77 \pm 0.63$ \\
    & Multisimilarity &  $84.33 \pm 0.21$ & $83.84 \pm 0.10$ & $82.03 \pm 0.38$ & $80.01 \pm 0.06$ & $77.14 \pm 0.22$ & $72.97 \pm 0.34$ & $69.78 \pm 0.37$ & $66.01 \pm 0.21$ \\
    & ArcFace &  $81.73 \pm 0.29$ & $81.66 \pm 0.39$ & $79.57 \pm 0.23$ & $77.19 \pm 0.06$ & $74.95 \pm 0.50$ & $69.35 \pm 0.26$ & $66.10 \pm 0.21$ & $62.55 \pm 0.62$ \\
    & ProxyAnchor &  $85.27 \pm 0.17$ & $84.81 \pm 0.05$ & $82.80 \pm 0.22$ & $80.23 \pm 0.19$ & $78.00 \pm 0.40$ & $73.09 \pm 0.21$ & $70.24 \pm 0.09$ & $66.35 \pm 0.17$ \\
    & R-Margin (D) &  $85.21 \pm 0.15$ & $84.46 \pm 0.34$ & $83.26 \pm 0.22$ & $80.73 \pm 0.28$ & $77.45 \pm 0.46$ & $74.42 \pm 0.07$ & $71.25 \pm 0.55$ & $69.20 \pm 0.28$ \\
    & Uniform Prior &  $84.56 \pm 0.16$ & $83.96 \pm 0.30$ & $82.20 \pm 0.15$ & $79.96 \pm 0.18$ & $77.48 \pm 0.37$ & $71.08 \pm 0.42$ & $69.06 \pm 0.64$ & $66.69 \pm 0.50$ \\
    & S2SD &  $87.93 \pm 0.07$ & $86.84 \pm 0.08$ & $85.59 \pm 0.10$ & $83.18 \pm 0.17$ & $80.55 \pm 0.16$ & $77.42 \pm 0.41$ & $74.64 \pm 0.16$ & $72.62 \pm 0.29$ \\
    & DiVA &  $86.74 \pm 0.08$ & $85.98 \pm 0.14$ & $84.43 \pm 0.11$ & $82.16 \pm 0.08$ & $79.28 \pm 0.35$ & $75.41 \pm 0.42$ & $72.76 \pm 0.38$ & $70.43 \pm 0.22$ \\
    
    \cmidrule[1pt]{1-10}                

    \multirow{8}{*}{mAP@1000} & Margin (D) &  $33.58 \pm 0.26$ & $33.23 \pm 0.01$ & $30.33 \pm 0.28$ & $28.50 \pm 0.28$ & $25.99 \pm 0.24$ & $21.43 \pm 0.17$ & $18.76 \pm 0.53$ & $16.56 \pm 0.22$ \\
    & Multisimilarity &  $34.01 \pm 0.29$ & $34.37 \pm 0.20$ & $31.39 \pm 0.28$ & $29.82 \pm 0.27$ & $28.09 \pm 0.16$ & $22.72 \pm 0.01$ & $20.39 \pm 0.09$ & $17.38 \pm 0.15$ \\
    & ArcFace &  $33.93 \pm 0.20$ & $34.19 \pm 0.22$ & $30.85 \pm 0.07$ & $28.51 \pm 0.28$ & $26.71 \pm 0.50$ & $20.67 \pm 0.24$ & $18.20 \pm 0.10$ & $15.40 \pm 0.29$ \\
    & ProxyAnchor &  $35.83 \pm 0.17$ & $36.22 \pm 0.20$ & $32.71 \pm 0.19$ & $31.07 \pm 0.24$ & $29.04 \pm 0.58$ & $23.08 \pm 0.08$ & $20.26 \pm 0.26$ & $17.16 \pm 0.10$ \\
    & R-Margin (D) &  $34.34 \pm 0.24$ & $34.69 \pm 0.20$ & $32.25 \pm 0.28$ & $30.63 \pm 0.29$ & $28.02 \pm 0.16$ & $23.12 \pm 0.30$ & $20.70 \pm 0.34$ & $18.68 \pm 0.06$ \\
    & Uniform Prior &  $34.03 \pm 0.23$ & $34.22 \pm 0.38$ & $31.05 \pm 0.52$ & $29.56 \pm 0.38$ & $26.99 \pm 0.25$ & $22.19 \pm 0.29$ & $20.06 \pm 0.46$ & $17.80 \pm 0.28$ \\
    & S2SD &  $37.41 \pm 0.14$ & $37.43 \pm 0.18$ & $34.48 \pm 0.15$ & $33.18 \pm 0.24$ & $30.93 \pm 0.32$ & $26.12 \pm 0.15$ & $23.56 \pm 0.25$ & $21.02 \pm 0.36$ \\
    & DiVA &  $36.60 \pm 0.40$ & $36.65 \pm 0.09$ & $32.96 \pm 0.40$ & $31.90 \pm 0.08$ & $29.40 \pm 0.24$ & $24.21 \pm 0.25$ & $22.10 \pm 0.24$ & $19.77 \pm 0.16$ \\

    \bottomrule
    \end{tabular}}
    \label{supp_tab:baseline_prog_full_car}
\end{table*}

\begin{table*}[t]
    \caption{DML generalization performance measured by Recall@1 and mAP@1000 on each train-test split of our \textit{ooDML} benchmark for the SOP dataset.}
  \centering
\resizebox{1\textwidth}{!}{
 \footnotesize
  \setlength\tabcolsep{1.4pt}
  %\begin{tabular}{l|c|ccc|c}
  \begin{tabular}{l|l||c|c|c|c|c|c|c|c}
     \toprule
     \textbf{Metric}  & \textbf{Method}$\downarrow$ | \textbf{Split (FID)}$\rightarrow$ & 1 (3.4) & 2 (24.6) & 3 (53.5) & 4 (99.4) & 5 (135.5) & 6 (155.3) & 7 (189.8) & 8 (235.1)\\
    \midrule
    \multirow{8}{*}{R@1} & Margin (D) &  $79.39 \pm 0.04$ & $78.58 \pm 0.05$ & $77.48 \pm 0.05$ & $76.18 \pm 0.05$ & $74.53 \pm 0.14$ & $72.62 \pm 0.05$ & $70.68 \pm 0.27$ & $69.71 \pm 0.09$ \\
    & Multisimilarity &  $79.31 \pm 0.09$ & $78.40 \pm 0.11$ & $77.40 \pm 0.06$ & $76.10 \pm 0.10$ & $74.45 \pm 0.07$ & $72.63 \pm 0.01$ & $70.90 \pm 0.10$ & $69.98 \pm 0.14$ \\
    & ArcFace &  $79.61 \pm 0.07$ & $78.46 \pm 0.10$ & $77.48 \pm 0.12$ & $75.67 \pm 0.03$ & $73.70 \pm 0.04$ & $71.78 \pm 0.04$ & $69.21 \pm 0.13$ & $67.89 \pm 0.14$ \\
    & ProxyAnchor &  $79.73 \pm 0.07$ & $78.60 \pm 0.02$ & $77.60 \pm 0.05$ & $75.70 \pm 0.05$ & $73.69 \pm 0.05$ & $71.96 \pm 0.12$ & $69.36 \pm 0.09$ & $68.07 \pm 0.03$ \\
    & R-Margin (D) &  $79.42 \pm 0.01$ & $78.50 \pm 0.01$ & $77.75 \pm 0.05$ & $76.44 \pm 0.07$ & $74.84 \pm 0.03$ & $73.15 \pm 0.00$ & $71.30 \pm 0.14$ & $70.21 \pm 0.16$ \\
    & Unifor mPrior &  $79.42 \pm 0.02$ & $78.61 \pm 0.04$ & $77.57 \pm 0.05$ & $76.12 \pm 0.07$ & $74.45 \pm 0.08$ & $72.68 \pm 0.06$ & $70.61 \pm 0.23$ & $69.65 \pm 0.12$ \\
    & S2SD &  $79.95 \pm 0.06$ & $78.88 \pm 0.09$ & $78.00 \pm 0.11$ & $76.66 \pm 0.15$ & $74.86 \pm 0.14$ & $73.33 \pm 0.06$ & $71.13 \pm 0.08$ & $70.19 \pm 0.09$ \\
    & DiVA &  $79.76 \pm 0.08$ & $78.75 \pm 0.05$ & $77.81 \pm 0.06$ & $76.58 \pm 0.05$ & $74.83 \pm 0.09$ & $73.15 \pm 0.03$ & $71.42 \pm 0.07$ & $70.30 \pm 0.15$ \\

    \cmidrule[1pt]{1-10}                

    \multirow{8}{*}{mAP@1000} & Margin (D) &  $47.47 \pm 0.03$ & $46.21 \pm 0.10$ & $45.16 \pm 0.09$ & $43.39 \pm 0.04$ & $41.79 \pm 0.18$ & $40.13 \pm 0.03$ & $37.64 \pm 0.31$ & $36.43 \pm 0.09$ \\
    & Multisimilarity &  $47.23 \pm 0.07$ & $45.85 \pm 0.09$ & $44.81 \pm 0.03$ & $43.17 \pm 0.03$ & $41.46 \pm 0.07$ & $39.67 \pm 0.10$ & $37.56 \pm 0.07$ & $36.76 \pm 0.13$ \\
    & ArcFace &  $47.76 \pm 0.05$ & $46.22 \pm 0.10$ & $45.09 \pm 0.05$ & $42.85 \pm 0.07$ & $40.86 \pm 0.10$ & $39.09 \pm 0.09$ & $36.01 \pm 0.11$ & $34.68 \pm 0.13$ \\
    & ProxyAnchor &  $47.57 \pm 0.04$ & $45.87 \pm 0.02$ & $44.89 \pm 0.03$ & $42.52 \pm 0.09$ & $40.67 \pm 0.01$ & $38.93 \pm 0.06$ & $35.90 \pm 0.05$ & $34.65 \pm 0.00$ \\
    & R-Margin (D) &  $47.94 \pm 0.04$ & $46.57 \pm 0.01$ & $45.59 \pm 0.05$ & $43.87 \pm 0.00$ & $42.26 \pm 0.11$ & $40.68 \pm 0.07$ & $38.44 \pm 0.19$ & $37.08 \pm 0.14$ \\
    & Uniform Prior &  $47.49 \pm 0.02$ & $46.28 \pm 0.07$ & $45.21 \pm 0.05$ & $43.39 \pm 0.05$ & $41.67 \pm 0.12$ & $40.07 \pm 0.01$ & $37.62 \pm 0.21$ & $36.37 \pm 0.10$ \\
    & S2SD &  $48.25 \pm 0.04$ & $47.08 \pm 0.18$ & $46.09 \pm 0.09$ & $44.27 \pm 0.07$ & $42.34 \pm 0.15$ & $40.87 \pm 0.11$ & $38.39 \pm 0.10$ & $36.98 \pm 0.11$ \\
    & DiVA &  $48.08 \pm 0.06$ & $46.57 \pm 0.05$ & $45.50 \pm 0.05$ & $43.77 \pm 0.04$ & $41.92 \pm 0.06$ & $40.44 \pm 0.10$ & $38.10 \pm 0.10$ & $36.73 \pm 0.07$ \\

    \bottomrule
    \end{tabular}}
    \label{supp_tab:baseline_prog_full_sop}
\end{table*}

\subsection{Influence of network capacity}
\begin{figure*}[t]
    \centering
    \includegraphics[width=0.95\textwidth]{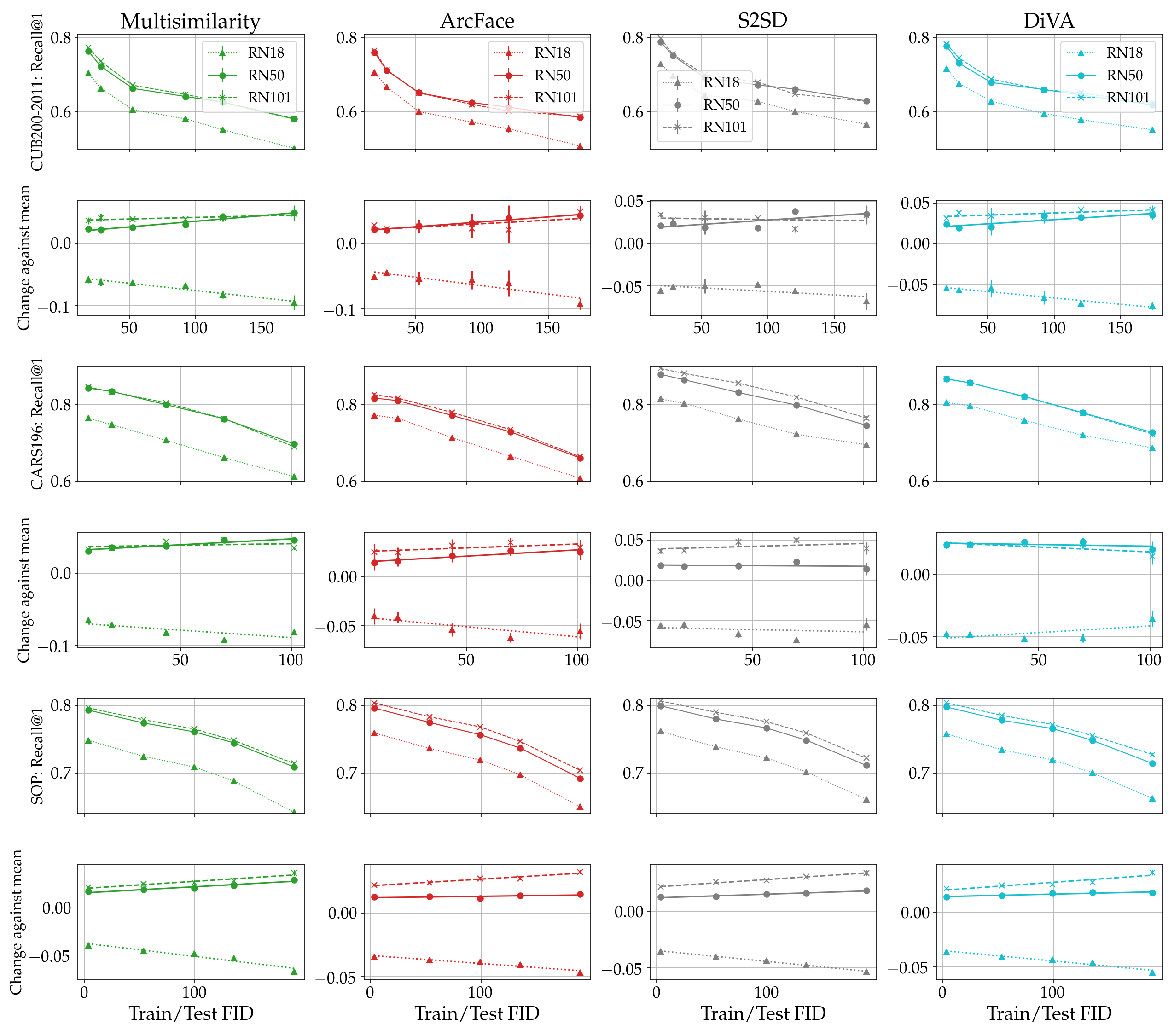}
    \caption{\textit{Generalization performance for different backbone architectures for varying distribution shifts on full ooDML benchmark (CUB200-2011, CARS196, SOP).} To reduce computational load, we only utilised two thirds of the studied splits. Overall, we show absolute Recall@1 performances averaged over 5 runs for each train-test split.}
    \label{supp_fig:arch_comp}
\end{figure*}

In Fig.~\ref{supp_fig:arch_comp} we present all results for our study on the influence of network capacity in Sec. \ref{sec:network_and_pretrain} in the main paper, in particular also for the remaining datasets CARS196 and SOP. 
Additionally, we show the differences in performances against the mean over all methods for each train-test split (Change against mean). As already discussed in Sec. \ref{sec:network_and_pretrain} in the main paper, these experiments similarly show that network capacity has only limited impact on OOD generalization, with benefits saturating eventually. 

\subsection{Measuring structural representation properties on ooDML}
\begin{figure}[h]
    \centering
    \includegraphics[width=1\textwidth]{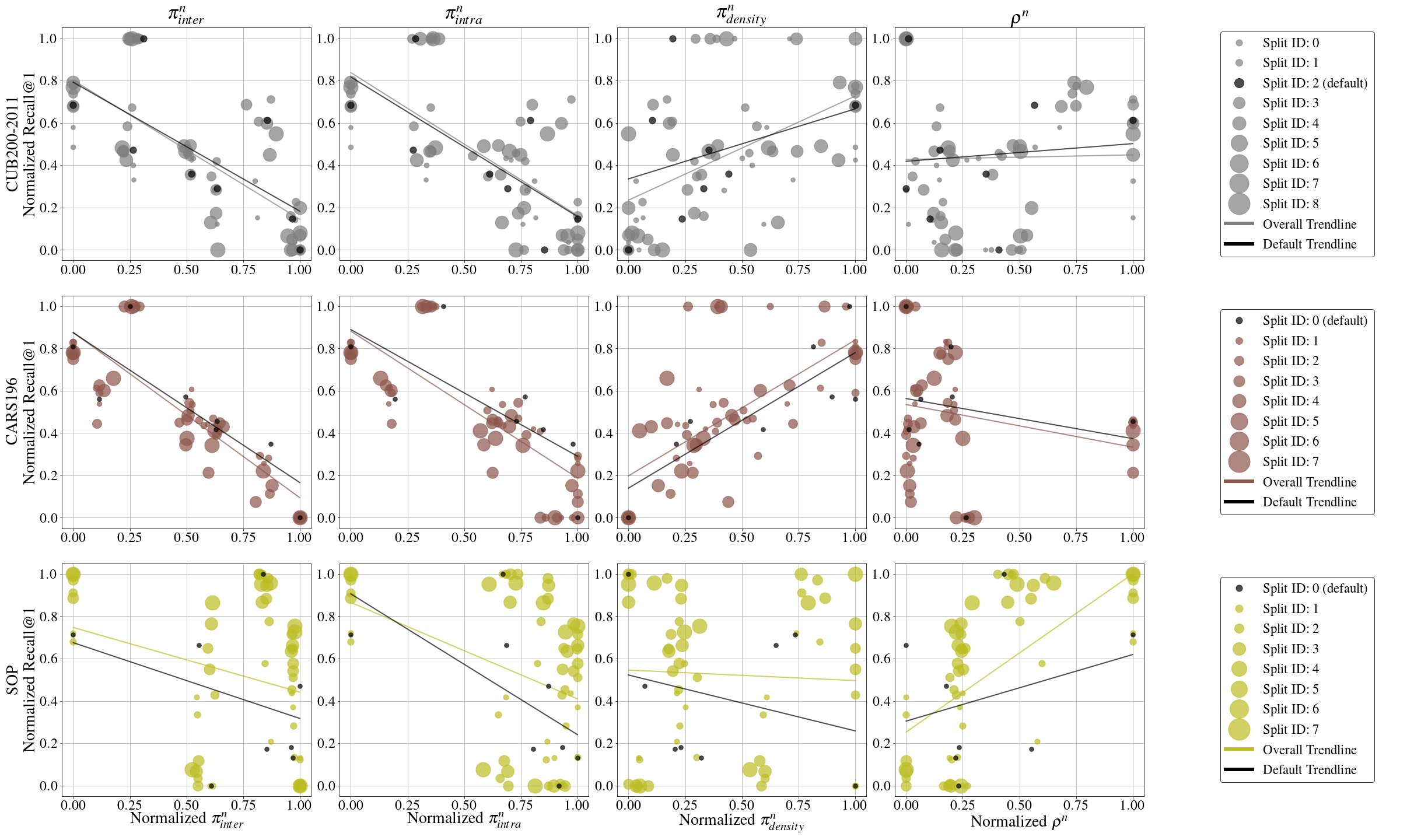}
    \caption{\textit{Generalization metrics computed on ooDML benchmark for all datasets} measured against Recall@1. Each column plots one of the (normalized) measured structural representation property (cf. Sec. 4.3 main paper) over the corresponding Recall@1 performance for all examined DML methods and distribution shifts. Trendlines are computed as least squares fit over all datapoints (overall), respectively only those corresponding to default splits (default).}
    \label{supp_fig:metrics_recall}
\end{figure}

\begin{figure}[h]
    \centering
    \includegraphics[width=1\textwidth]{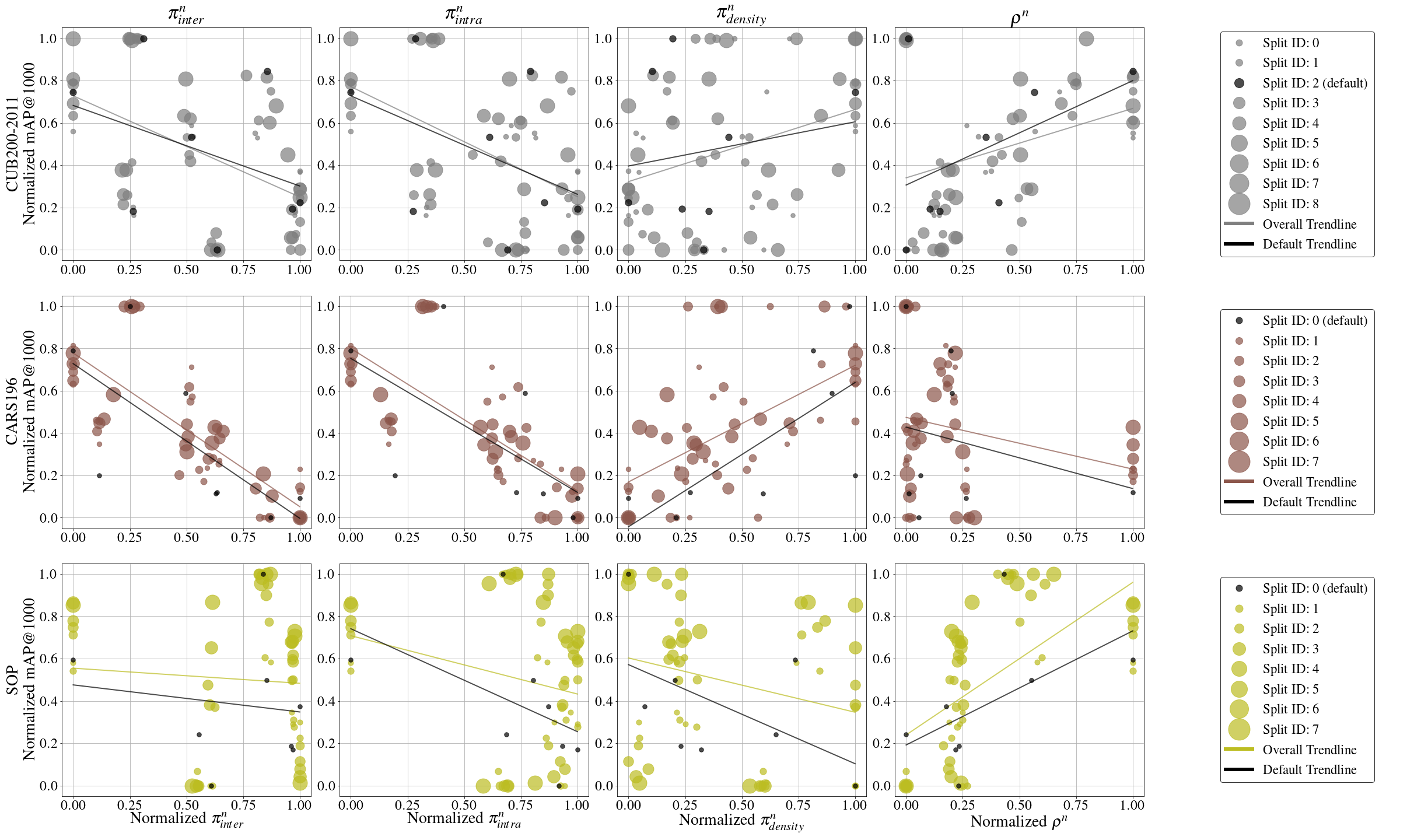}
    \caption{\textit{Generalization metrics computed on ooDML benchmark for all datasets} measured against mAP@1000. Each column plots one of the (normalized) measured structural representation property (cf. Sec. 4.3 main paper) over the corresponding mAP@1000 performance for all examined DML methods and distribution shifts. Trendlines are computed as least squares fit over all datapoints (overall), respectively only those corresponding to default splits (default).}
    \label{supp_fig:metrics_map}
\end{figure}

This section extends the results presented in Sec. \ref{sec:gen_metrics}.
We show results for all datasets, i.e. CUB200-2011, CARS196 and SOP, for all metrics measuring structural representation properties discussed in Sec. \ref{sec:gen_metrics} in the main paper. 
We analyze correlations of these metrics with generalization performance both based on Recall@1 (Fig.~\ref{supp_fig:metrics_recall}) and mAP@1000 (Fig.~\ref{supp_fig:metrics_map}). As discussed in the main paper, independent of the underlying performance metric, none of the structural representation properties show consistent correlation with generalization performance across all datasets, suggesting further research into meaningful latent space properties that can be linked to zero-shot generalization independent of chosen objectives and shifts.

\subsection{Few-Shot DML}
\begin{figure}[t]
\begin{subfigure}[t]{1\textwidth}
    \centering
    \includegraphics[width=1\textwidth]{images/few_shot_compare_cub_R1.png}
    \caption{}
    \label{supp_fig:fewshot_cub_recall}
\end{subfigure}
\begin{subfigure}[t]{1\textwidth}
    \centering
    \includegraphics[width=1\textwidth]{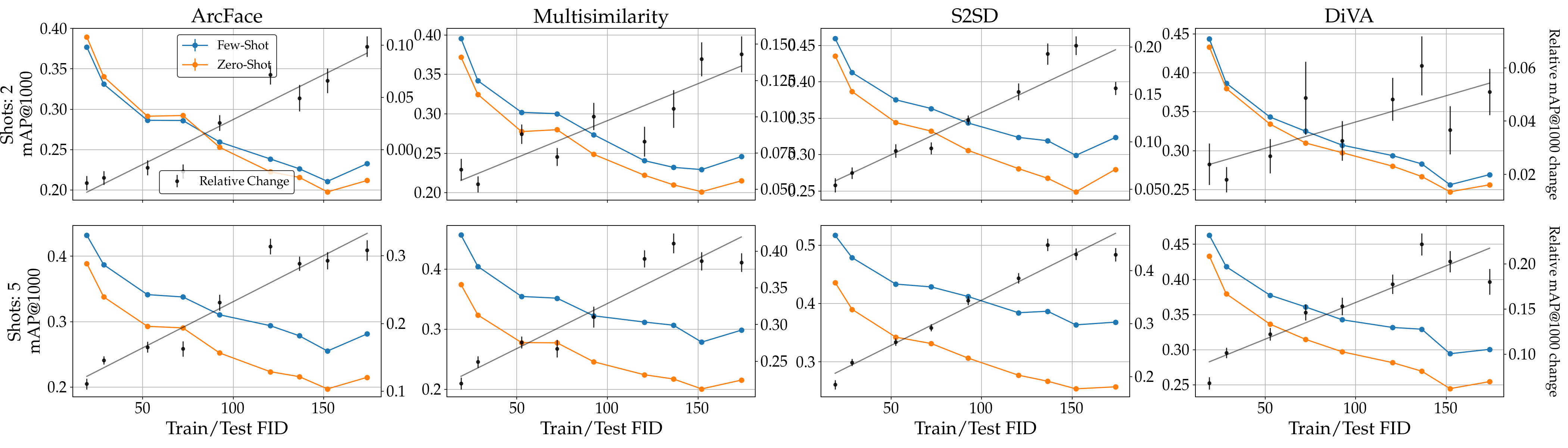}
    \caption{}
    \label{supp_fig:fewshot_cub_map}
\end{subfigure}
\begin{subfigure}[t]{1\textwidth}
    \centering
    \includegraphics[width=1\textwidth]{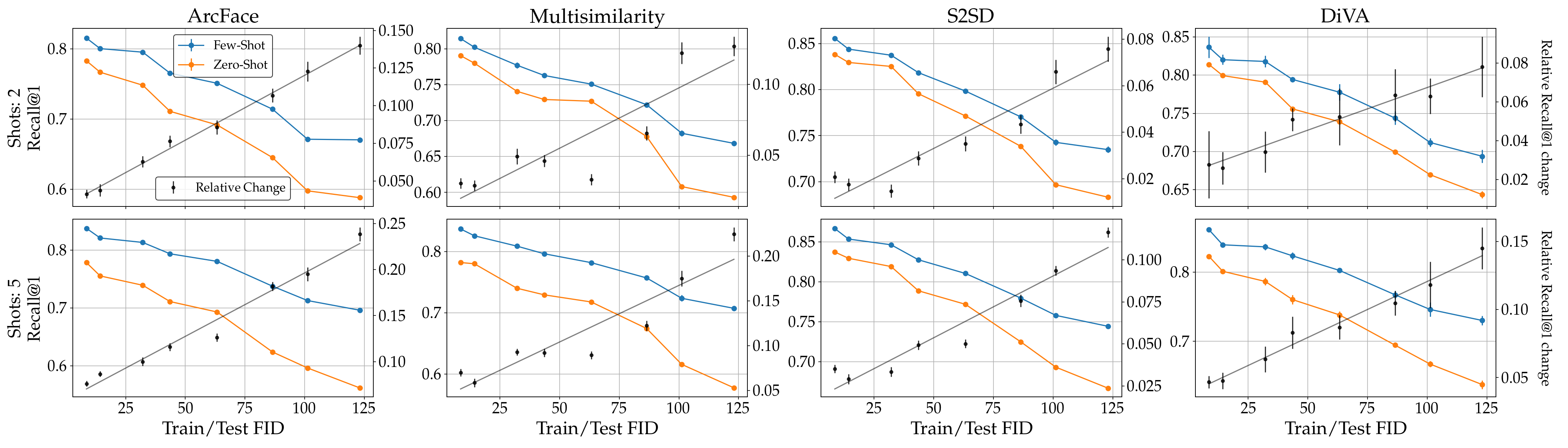}
    \caption{}
    \label{supp_fig:fewshot_car_recall}
\end{subfigure}
\begin{subfigure}[t]{1\textwidth}
    \centering
    \includegraphics[width=1\textwidth]{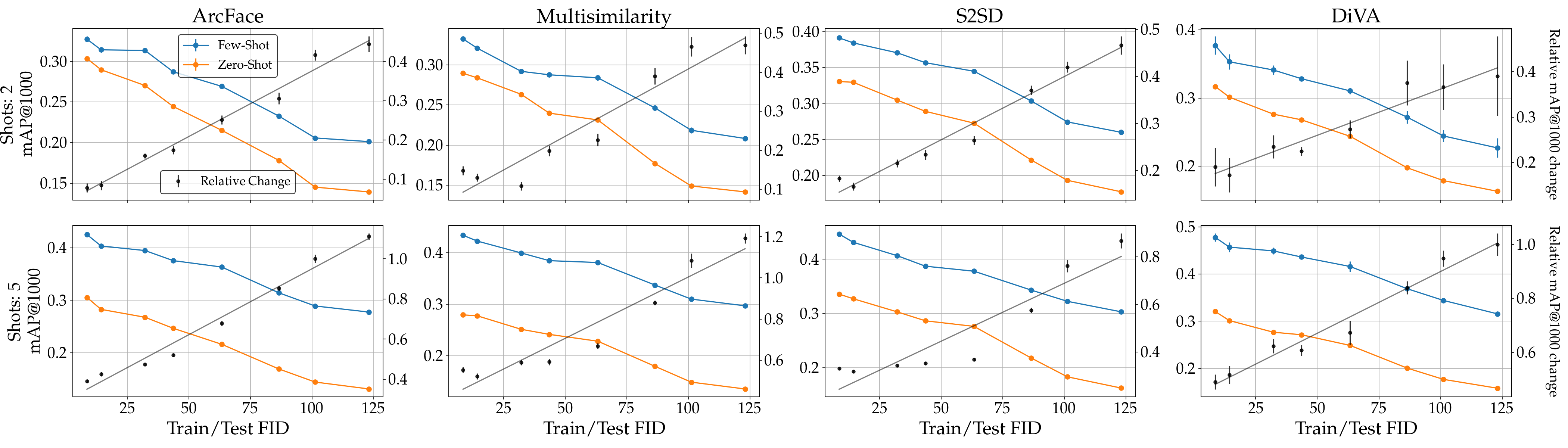}
    \caption{}
    \label{supp_fig:fewshot_car_map}
\end{subfigure}
\caption{\textit{Few-Shot adaptation of DML representations on CUB200-2011 and CARS196.} Columns show average Recall@1 performance over 10 episodes of 2- and 5-shot adaption as well as the baseline zero-shot DML results on the same train-test splits (based on \textit{ooDML} benchmark) for various DML approaches (\blue{fewshot} and \orange{zeroshot}), highlighting a substantial benefit of few-shot adaptation for \textit{a priori} unknown distribution shifts (see black line highlighting relative improvements). Relative improvements are computed as relative change of few-shot performance against respective zero-shot performance.}
\label{supp_fig:fewshot_all}
\end{figure}

In Sec. \ref{subsec:fewshot} in the main paper, we analyzed few-shot adaption of DML representations to novel test distributions as a remedy to bridge their distribution shift to the training data. 
This section extends showcased results: Fig.~\ref{supp_fig:fewshot_all} presents all our results on both CUB200-2011 (a+b) and CARS196 (c+d) dataset based on both Recall@1 and mAP@1000. 
The results on CARS196 verify the consistent improvement of leveraging very few examples for embedding space adaption over strict zero-shot transfer based on the original DML representation that we already observed for the CUB200-2011 dataset, which holds disproportionally well for larger distribution shifts. The corresponding data basis for Fig.~\ref{supp_fig:fewshot_all} is presented in Tab.~\ref{supp_tab:few_shot_cub} for the CUB200-2011 dataset and in Tab.~\ref{supp_tab:few_shot_car} for the CARS196 dataset.
%\\
%Finally, Fig.~\ref{fig:pretrained_fewshot_comp} repeats the experiment 'Generic representations versus Deep Metric Learning' in Sec. 4.4 in the main paper based on the few-shot adapted representations.... is this really needed ?

\begin{table*}[t]
    \caption{Evaluation of zero-generalization and subsequent few-shot adaptation measured by Recall@1 and mAP@1000 based on few-shot dataplits built from the train-test splits of the \textit{ooDML} benchmark (CUB200-2011). Results are further summarized in Fig.~\ref{supp_fig:fewshot_all} (a) and (b).}
  \centering
\resizebox{1\textwidth}{!}{
 \footnotesize
  \setlength\tabcolsep{1.4pt}
  %\begin{tabular}{l|c|ccc|c}
  \begin{tabular}{l||c|c|l||c|c|c|c|c|c|c|c|c}
     \toprule
     \textbf{Metric} & \textbf{Shot} & \textbf{Use} & \textbf{Method}$\downarrow$ | \textbf{Split}$\rightarrow$ & 1 & 2 & 3 & 4 & 5 & 6 & 7 & 8 & 9\\
    \midrule
    \multirow{16}{*}[-8pt]{R@1} & \multirow{8}{*}{2} & \multirow{4}{*}{Zero} & ArcFace & $69.88 \pm 0.12$ & $65.06 \pm 0.12$ & $58.76 \pm 0.12$ & $58.02 \pm 0.10$ & $55.14 \pm 0.14$ & $54.43 \pm 0.11$ & $52.04 \pm 0.12$ & $49.49 \pm 0.13$ & $51.00 \pm 0.21$\\
    &&& Multisimilarity & $70.02 \pm 0.09$ & $64.87 \pm 0.11$ & $59.88 \pm 0.11$ & $60.71 \pm 0.17$ & $56.96 \pm 0.16$ & $54.82 \pm 0.18$ & $52.33 \pm 0.11$ & $51.67 \pm 0.14$ & $52.45 \pm 0.13$\\
    &&& S2SD & $75.35 \pm 0.15$ & $71.38 \pm 0.14$ & $66.56 \pm 0.14$ & $65.85 \pm 0.10$ & $64.47 \pm 0.14$ & $63.56 \pm 0.13$ & $61.66 \pm 0.12$ & $58.87 \pm 0.17$ & $60.12 \pm 0.10$\\
    &&& DiVA & $74.86 \pm 0.10$ & $70.68 \pm 0.16$ & $65.17 \pm 0.12$ & $63.83 \pm 0.19$ & $62.63 \pm 0.12$ & $62.48 \pm 0.10$ & $60.45 \pm 0.12$ & $57.99 \pm 0.16$ & $57.96 \pm 0.16$\\
    
    \cmidrule[0.5pt]{3-13}                
    && \multirow{4}{*}{Few} & ArcFace & $70.88 \pm 0.31$ & $66.76 \pm 0.39$ & $60.97 \pm 0.22$ & $60.77 \pm 0.35$ & $58.45 \pm 0.39$ & $57.42 \pm 0.34$ & $55.55 \pm 0.49$ & $52.11 \pm 0.33$ & $55.01 \pm 0.55$\\
    &&& Multisimilarity & $72.26 \pm 0.32$ & $67.96 \pm 0.29$ & $62.64 \pm 0.27$ & $62.82 \pm 0.29$ & $60.52 \pm 0.28$ & $58.90 \pm 0.32$ & $56.82 \pm 0.31$ & $56.07 \pm 0.37$ & $56.92 \pm 0.39$\\
    &&& S2SD & $76.03 \pm 0.33$ & $71.90 \pm 0.34$ & $67.56 \pm 0.36$ & $66.81 \pm 0.31$ & $66.22 \pm 0.27$ & $65.45 \pm 0.27$ & $64.34 \pm 0.40$ & $61.26 \pm 0.33$ & $62.96 \pm 0.33$\\
    &&& DiVA & $75.07 \pm 0.24$ & $70.96 \pm 0.43$ & $65.96 \pm 0.34$ & $64.38 \pm 0.40$ & $63.79 \pm 0.35$ & $63.26 \pm 0.32$ & $62.30 \pm 0.29$ & $59.49 \pm 0.36$ & $59.54 \pm 0.44$\\
    
    \cmidrule[1pt]{2-13}   
    & \multirow{8}{*}{5} & \multirow{4}{*}{Zero} & ArcFace & $69.96 \pm 0.15$ & $64.44 \pm 0.12$ & $58.43 \pm 0.17$ & $57.67 \pm 0.21$ & $54.42 \pm 0.27$ & $54.01 \pm 0.28$ & $51.85 \pm 0.28$ & $49.08 \pm 0.26$ & $50.61 \pm 0.21$\\
    &&& Multisimilarity & $70.15 \pm 0.19$ & $65.66 \pm 0.19$ & $59.42 \pm 0.27$ & $59.71 \pm 0.13$ & $56.59 \pm 0.18$ & $55.39 \pm 0.19$ & $54.26 \pm 0.31$ & $51.20 \pm 0.21$ & $52.10 \pm 0.21$\\
    &&& S2SD & $75.24 \pm 0.22$ & $71.81 \pm 0.10$ & $65.88 \pm 0.22$ & $65.03 \pm 0.24$ & $63.49 \pm 0.27$ & $62.69 \pm 0.30$ & $61.81 \pm 0.22$ & $57.40 \pm 0.19$ & $57.55 \pm 0.18$\\
    &&& DiVA & $74.71 \pm 0.18$ & $70.37 \pm 0.17$ & $65.18 \pm 0.15$ & $64.93 \pm 0.11$ & $62.27 \pm 0.21$ & $62.17 \pm 0.22$ & $60.40 \pm 0.29$ & $57.42 \pm 0.26$ & $57.26 \pm 0.26$\\
    
    \cmidrule[0.5pt]{3-13}       
    & & \multirow{4}{*}{Few} & ArcFace & $73.41 \pm 0.31$ & $69.19 \pm 0.32$ & $64.16 \pm 0.53$ & $63.87 \pm 0.34$ & $61.92 \pm 0.43$ & $61.48 \pm 0.33$ & $59.86 \pm 0.40$ & $55.89 \pm 0.49$ & $58.98 \pm 0.53$\\
    &&& Multisimilarity & $74.66 \pm 0.33$ & $70.88 \pm 0.26$ & $65.95 \pm 0.38$ & $65.62 \pm 0.35$ & $63.36 \pm 0.36$ & $63.55 \pm 0.35$ & $62.40 \pm 0.35$ & $59.11 \pm 0.45$ & $60.88 \pm 0.37$\\
    &&& S2SD & $75.70 \pm 0.21$ & $72.88 \pm 0.35$ & $68.77 \pm 0.41$ & $68.46 \pm 0.23$ & $67.75 \pm 0.29$ & $66.38 \pm 0.41$ & $66.24 \pm 0.54$ & $63.37 \pm 0.51$ & $63.56 \pm 0.43$\\
    &&& DiVA & $76.05 \pm 0.40$ & $72.61 \pm 0.25$ & $67.93 \pm 0.40$ & $67.42 \pm 0.46$ & $65.59 \pm 0.38$ & $65.43 \pm 0.38$ & $65.02 \pm 0.39$ & $61.21 \pm 0.45$ & $61.28 \pm 0.41$\\
    
    \cmidrule[1.25pt]{1-13}    
    \multirow{16}{*}[-9pt]{mAP@1000} & \multirow{8}{*}{2} & \multirow{4}{*}{Zero} & ArcFace & $38.95 \pm 0.08$ & $34.03 \pm 0.06$ & $29.13 \pm 0.04$ & $29.22 \pm 0.05$ & $25.29 \pm 0.04$ & $22.25 \pm 0.05$ & $21.56 \pm 0.05$ & $19.76 \pm 0.04$ & $21.17 \pm 0.05$\\
    &&& Multisimilarity & $37.19 \pm 0.07$ & $32.46 \pm 0.07$ & $27.75 \pm 0.04$ & $27.99 \pm 0.04$ & $24.86 \pm 0.03$ & $22.22 \pm 0.10$ & $21.00 \pm 0.06$ & $20.12 \pm 0.06$ & $21.51 \pm 0.05$\\
    &&& S2SD & $43.56 \pm 0.09$ & $38.66 \pm 0.07$ & $34.41 \pm 0.05$ & $33.23 \pm 0.05$ & $30.59 \pm 0.04$ & $28.08 \pm 0.05$ & $26.76 \pm 0.06$ & $24.89 \pm 0.04$ & $27.98 \pm 0.04$\\
    &&& DiVA & $43.33 \pm 0.08$ & $37.99 \pm 0.05$ & $33.43 \pm 0.04$ & $31.00 \pm 0.05$ & $29.75 \pm 0.03$ & $28.02 \pm 0.07$ & $26.69 \pm 0.06$ & $24.73 \pm 0.05$ & $25.63 \pm 0.05$\\
    
    \cmidrule[0.5pt]{3-13}    
    & & \multirow{4}{*}{Few} & ArcFace & $37.71 \pm 0.25$ & $33.11 \pm 0.21$ & $28.63 \pm 0.21$ & $28.60 \pm 0.19$ & $25.94 \pm 0.18$ & $23.84 \pm 0.20$ & $22.62 \pm 0.27$ & $21.06 \pm 0.23$ & $23.26 \pm 0.20$\\
    &&& Multisimilarity & $39.57 \pm 0.27$ & $34.20 \pm 0.16$ & $30.20 \pm 0.18$ & $30.02 \pm 0.17$ & $27.36 \pm 0.23$ & $24.07 \pm 0.20$ & $23.22 \pm 0.27$ & $22.94 \pm 0.23$ & $24.59 \pm 0.26$\\
    &&& S2SD & $45.94 \pm 0.31$ & $41.26 \pm 0.22$ & $37.53 \pm 0.23$ & $36.33 \pm 0.20$ & $34.35 \pm 0.14$ & $32.37 \pm 0.24$ & $31.91 \pm 0.29$ & $29.90 \pm 0.24$ & $32.36 \pm 0.19$\\
    &&& DiVA & $44.36 \pm 0.33$ & $38.67 \pm 0.17$ & $34.33 \pm 0.21$ & $32.51 \pm 0.42$ & $30.72 \pm 0.22$ & $29.37 \pm 0.21$ & $28.31 \pm 0.29$ & $25.63 \pm 0.22$ & $26.94 \pm 0.22$\\
    
    \cmidrule[1pt]{2-13}                
    & \multirow{8}{*}{5} & \multirow{4}{*}{Zero} & ArcFace & $38.91 \pm 0.10$ & $33.79 \pm 0.05$ & $29.30 \pm 0.06$ & $29.06 \pm 0.08$ & $25.24 \pm 0.07$ & $22.38 \pm 0.09$ & $21.62 \pm 0.09$ & $19.75 \pm 0.05$ & $21.52 \pm 0.06$\\
    &&& Multisimilarity & $37.45 \pm 0.09$ & $32.38 \pm 0.06$ & $27.80 \pm 0.06$ & $27.76 \pm 0.09$ & $24.61 \pm 0.07$ & $22.46 \pm 0.08$ & $21.75 \pm 0.10$ & $20.10 \pm 0.06$ & $21.56 \pm 0.07$\\
    &&& S2SD & $43.61 \pm 0.12$ & $39.01 \pm 0.08$ & $34.25 \pm 0.07$ & $33.16 \pm 0.09$ & $30.67 \pm 0.07$ & $27.73 \pm 0.09$ & $26.70 \pm 0.12$ & $25.40 \pm 0.07$ & $25.74 \pm 0.12$\\
    &&& DiVA & $43.34 \pm 0.10$ & $37.96 \pm 0.08$ & $33.65 \pm 0.07$ & $31.49 \pm 0.12$ & $29.72 \pm 0.08$ & $28.17 \pm 0.10$ & $26.95 \pm 0.12$ & $24.47 \pm 0.09$ & $25.47 \pm 0.09$\\
    
    \cmidrule[0.5pt]{3-13}      
    & & \multirow{4}{*}{Few} & ArcFace & $43.22 \pm 0.28$ & $38.71 \pm 0.20$ & $34.14 \pm 0.22$ & $33.79 \pm 0.32$ & $31.06 \pm 0.29$ & $29.41 \pm 0.23$ & $27.86 \pm 0.20$ & $25.53 \pm 0.25$ & $28.15 \pm 0.32$\\
    &&& Multisimilarity & $45.68 \pm 0.28$ & $40.45 \pm 0.24$ & $35.46 \pm 0.21$ & $35.18 \pm 0.31$ & $32.25 \pm 0.34$ & $31.21 \pm 0.24$ & $30.69 \pm 0.26$ & $27.88 \pm 0.24$ & $29.86 \pm 0.26$\\
    &&& S2SD & $51.67 \pm 0.35$ & $47.86 \pm 0.25$ & $43.33 \pm 0.22$ & $42.85 \pm 0.19$ & $41.21 \pm 0.24$ & $38.43 \pm 0.25$ & $38.68 \pm 0.26$ & $36.35 \pm 0.26$ & $36.82 \pm 0.28$\\
    &&& DiVA & $46.29 \pm 0.27$ & $41.82 \pm 0.20$ & $37.77 \pm 0.22$ & $36.10 \pm 0.24$ & $34.27 \pm 0.28$ & $33.17 \pm 0.28$ & $32.93 \pm 0.29$ & $29.44 \pm 0.27$ & $30.07 \pm 0.35$\\

    \bottomrule
    \end{tabular}}
    \label{supp_tab:few_shot_cub}
\end{table*}

\begin{table*}[t]
    \caption{Evaluation of zero-generalization and subsequent few-shot adaptation measured by Recall@1 and mAP@1000 based on few-shot dataplits built from the train-test splits of the \textit{ooDML} benchmark (CARS196). Results are further summarized in Fig.~\ref{supp_fig:fewshot_all} (c) and (d).}
  \centering
\resizebox{1\textwidth}{!}{
 \footnotesize
  \setlength\tabcolsep{1.4pt}
  %\begin{tabular}{l|c|ccc|c}
  \begin{tabular}{l||c|c|l||c|c|c|c|c|c|c|c|c}
     \toprule
     \textbf{Metric} & \textbf{Shot} & \textbf{Use} & \textbf{Method}$\downarrow$ | \textbf{Split}$\rightarrow$ & 1 & 2 & 3 & 4 & 5 & 6 & 7 & 8\\
    \midrule
\multirow{16}{*}[-9pt]{R@1} & \multirow{8}{*}{2} & \multirow{4}{*}{Zero} & ArcFace & $78.30 \pm 0.08$ & $76.67 \pm 0.09$ & $74.82 \pm 0.08$ & $71.11 \pm 0.08$ & $69.18 \pm 0.08$ & $64.53 \pm 0.12$ & $59.79 \pm 0.09$ & $58.82 \pm 0.09$\\
&&& Multisimilarity & $79.07 \pm 0.07$ & $78.02 \pm 0.09$ & $74.10 \pm 0.06$ & $72.96 \pm 0.11$ & $72.72 \pm 0.06$ & $67.80 \pm 0.17$ & $60.82 \pm 0.14$ & $59.31 \pm 0.13$\\
&&& S2SD & $83.84 \pm 0.06$ & $82.96 \pm 0.09$ & $82.53 \pm 0.08$ & $79.56 \pm 0.07$ & $77.14 \pm 0.10$ & $73.84 \pm 0.09$ & $69.67 \pm 0.13$ & $68.31 \pm 0.11$\\
&&& DiVA & $81.39 \pm 0.20$ & $79.94 \pm 0.08$ & $79.08 \pm 0.35$ & $75.58 \pm 0.28$ & $73.89 \pm 0.18$ & $69.94 \pm 0.31$ & $66.97 \pm 0.23$ & $64.36 \pm 0.48$\\

\cmidrule[0.5pt]{3-12}                
&& \multirow{4}{*}{Few} & ArcFace & $81.54 \pm 0.21$ & $80.03 \pm 0.30$ & $79.53 \pm 0.26$ & $76.54 \pm 0.26$ & $75.10 \pm 0.31$ & $71.42 \pm 0.26$ & $67.13 \pm 0.38$ & $67.05 \pm 0.34$\\
&&& Multisimilarity & $81.45 \pm 0.28$ & $80.25 \pm 0.27$ & $77.74 \pm 0.41$ & $76.31 \pm 0.29$ & $75.10 \pm 0.28$ & $72.23 \pm 0.29$ & $68.25 \pm 0.44$ & $66.84 \pm 0.38$\\
&&& S2SD & $85.57 \pm 0.20$ & $84.41 \pm 0.20$ & $83.74 \pm 0.21$ & $81.85 \pm 0.23$ & $79.85 \pm 0.23$ & $77.04 \pm 0.28$ & $74.27 \pm 0.34$ & $73.48 \pm 0.35$\\
&&& DiVA & $83.65 \pm 1.39$ & $82.02 \pm 0.66$ & $81.79 \pm 0.75$ & $79.43 \pm 0.33$ & $77.76 \pm 1.06$ & $74.38 \pm 0.88$ & $71.19 \pm 0.57$ & $69.38 \pm 0.86$\\

\cmidrule[1pt]{2-12}   
& \multirow{8}{*}{5} & \multirow{4}{*}{Zero} & ArcFace & $77.82 \pm 0.14$ & $75.52 \pm 0.11$ & $73.91 \pm 0.12$ & $71.10 \pm 0.16$ & $69.30 \pm 0.16$ & $62.40 \pm 0.12$ & $59.64 \pm 0.15$ & $56.20 \pm 0.18$\\
&&& Multisimilarity & $78.19 \pm 0.15$ & $78.00 \pm 0.14$ & $74.00 \pm 0.18$ & $72.93 \pm 0.16$ & $71.77 \pm 0.17$ & $67.44 \pm 0.19$ & $61.59 \pm 0.15$ & $57.77 \pm 0.18$\\
&&& S2SD & $83.75 \pm 0.11$ & $82.96 \pm 0.13$ & $81.92 \pm 0.13$ & $78.88 \pm 0.12$ & $77.19 \pm 0.09$ & $72.47 \pm 0.19$ & $69.30 \pm 0.10$ & $66.66 \pm 0.09$\\
&&& DiVA & $82.27 \pm 0.10$ & $80.12 \pm 0.31$ & $78.65 \pm 0.56$ & $76.03 \pm 0.67$ & $73.82 \pm 0.53$ & $69.46 \pm 0.26$ & $66.72 \pm 0.43$ & $63.76 \pm 0.63$\\

\cmidrule[0.5pt]{3-12}  
&& \multirow{4}{*}{Few} & ArcFace & $83.74 \pm 0.19$ & $82.07 \pm 0.19$ & $81.30 \pm 0.33$ & $79.34 \pm 0.25$ & $78.04 \pm 0.25$ & $73.75 \pm 0.26$ & $71.27 \pm 0.40$ & $69.60 \pm 0.36$\\
&&& Multisimilarity & $83.67 \pm 0.28$ & $82.56 \pm 0.35$ & $80.87 \pm 0.20$ & $79.63 \pm 0.29$ & $78.17 \pm 0.26$ & $75.68 \pm 0.32$ & $72.37 \pm 0.52$ & $70.73 \pm 0.39$\\
&&& S2SD & $86.69 \pm 0.17$ & $85.37 \pm 0.21$ & $84.65 \pm 0.19$ & $82.77 \pm 0.18$ & $81.05 \pm 0.18$ & $77.96 \pm 0.18$ & $75.79 \pm 0.17$ & $74.43 \pm 0.18$\\
&&& DiVA & $86.11 \pm 0.36$ & $83.94 \pm 0.35$ & $83.64 \pm 0.45$ & $82.35 \pm 0.52$ & $80.24 \pm 0.31$ & $76.73 \pm 0.56$ & $74.61 \pm 1.02$ & $73.02 \pm 0.67$\\

\cmidrule[1.25pt]{1-12}  
\multirow{16}{*}[-9pt]{mAP@1000} & \multirow{8}{*}{2} & \multirow{4}{*}{Zero} & ArcFace & $30.37 \pm 0.04$ & $28.98 \pm 0.03$ & $27.03 \pm 0.02$ & $24.46 \pm 0.03$ & $21.52 \pm 0.03$ & $17.81 \pm 0.05$ & $14.52 \pm 0.03$ & $13.93 \pm 0.03$\\
&&& Multisimilarity & $28.95 \pm 0.03$ & $28.38 \pm 0.02$ & $26.33 \pm 0.02$ & $23.98 \pm 0.04$ & $23.13 \pm 0.03$ & $17.71 \pm 0.05$ & $14.90 \pm 0.03$ & $14.17 \pm 0.04$\\
&&& S2SD & $33.08 \pm 0.04$ & $32.96 \pm 0.04$ & $30.49 \pm 0.02$ & $28.92 \pm 0.05$ & $27.28 \pm 0.03$ & $22.16 \pm 0.05$ & $19.33 \pm 0.04$ & $17.74 \pm 0.05$\\
&&& DiVA & $31.69 \pm 0.11$ & $30.13 \pm 0.08$ & $27.64 \pm 0.03$ & $26.80 \pm 0.10$ & $24.38 \pm 0.12$ & $19.77 \pm 0.22$ & $17.89 \pm 0.12$ & $16.32 \pm 0.13$\\

\cmidrule[0.5pt]{3-12}  
&& \multirow{4}{*}{Few} & ArcFace & $32.75 \pm 0.35$ & $31.43 \pm 0.35$ & $31.35 \pm 0.16$ & $28.74 \pm 0.27$ & $26.93 \pm 0.24$ & $23.27 \pm 0.25$ & $20.58 \pm 0.19$ & $20.12 \pm 0.28$\\
&&& Multisimilarity & $33.23 \pm 0.32$ & $32.08 \pm 0.29$ & $29.19 \pm 0.29$ & $28.75 \pm 0.32$ & $28.38 \pm 0.36$ & $24.63 \pm 0.36$ & $21.85 \pm 0.36$ & $20.82 \pm 0.32$\\
&&& S2SD & $39.14 \pm 0.22$ & $38.44 \pm 0.27$ & $37.06 \pm 0.25$ & $35.69 \pm 0.31$ & $34.50 \pm 0.25$ & $30.38 \pm 0.20$ & $27.46 \pm 0.22$ & $26.01 \pm 0.33$\\
&&& DiVA & $37.71 \pm 1.35$ & $35.32 \pm 1.14$ & $34.11 \pm 0.71$ & $32.83 \pm 0.23$ & $31.05 \pm 0.46$ & $27.19 \pm 0.93$ & $24.44 \pm 0.88$ & $22.68 \pm 1.42$\\

\cmidrule[1pt]{2-12}  
& \multirow{8}{*}{5} & \multirow{4}{*}{Zero} & ArcFace & $30.55 \pm 0.08$ & $28.28 \pm 0.06$ & $26.78 \pm 0.05$ & $24.68 \pm 0.06$ & $21.64 \pm 0.04$ & $16.94 \pm 0.05$ & $14.46 \pm 0.04$ & $13.14 \pm 0.06$\\
&&& Multisimilarity & $27.95 \pm 0.07$ & $27.78 \pm 0.07$ & $25.15 \pm 0.06$ & $24.16 \pm 0.07$ & $22.83 \pm 0.04$ & $17.95 \pm 0.04$ & $14.88 \pm 0.05$ & $13.55 \pm 0.05$\\
&&& S2SD & $33.57 \pm 0.06$ & $32.71 \pm 0.09$ & $30.29 \pm 0.07$ & $28.64 \pm 0.06$ & $27.61 \pm 0.05$ & $21.78 \pm 0.05$ & $18.30 \pm 0.19$ & $16.22 \pm 0.22$\\
&&& DiVA & $32.06 \pm 0.13$ & $30.14 \pm 0.10$ & $27.67 \pm 0.12$ & $27.13 \pm 0.22$ & $24.87 \pm 0.07$ & $20.05 \pm 0.11$ & $17.66 \pm 0.19$ & $15.79 \pm 0.22$\\

\cmidrule[0.5pt]{3-12}  
&& \multirow{4}{*}{Few} & ArcFace & $42.52 \pm 0.21$ & $40.33 \pm 0.31$ & $39.46 \pm 0.17$ & $37.55 \pm 0.21$ & $36.33 \pm 0.28$ & $31.42 \pm 0.20$ & $28.92 \pm 0.28$ & $27.76 \pm 0.16$\\
&&& Multisimilarity & $43.43 \pm 0.37$ & $42.30 \pm 0.38$ & $39.99 \pm 0.33$ & $38.49 \pm 0.35$ & $38.12 \pm 0.31$ & $33.75 \pm 0.19$ & $31.02 \pm 0.51$ & $29.71 \pm 0.31$\\
&&& S2SD & $44.67 \pm 0.12$ & $43.13 \pm 0.16$ & $40.66 \pm 0.15$ & $38.72 \pm 0.17$ & $37.80 \pm 0.14$ & $34.31 \pm 0.24$ & $32.24 \pm 0.33$ & $30.30 \pm 0.30$\\
&&& DiVA & $47.79 \pm 0.85$ & $45.72 \pm 1.00$ & $44.92 \pm 0.72$ & $43.62 \pm 0.42$ & $41.62 \pm 1.09$ & $36.89 \pm 0.43$ & $34.41 \pm 0.37$ & $31.56 \pm 0.49$\\

\bottomrule
\end{tabular}}
\label{supp_tab:few_shot_car}
\end{table*}

%\input{figures/pretrained_fewshot_comp}

% %%%%%%%%%%%%%%%%%%%%%%%%%%%%%%%%%%
% \clearpage

% {\small
% \bibliographystyle{ieee_fullname.bst}
% \bibliography{egbib}
% }
% \end{document}

\end{document}